\documentclass{article}
\usepackage{arxiv}

\usepackage[utf8]{inputenc} 
\usepackage[T1]{fontenc}    
\usepackage{hyperref}       
\usepackage{url}            
\usepackage{booktabs}       
\usepackage{amsfonts}       
\usepackage{nicefrac}       
\usepackage{microtype}      
\usepackage{lipsum}         
\usepackage{graphicx}
\usepackage{doi}

\usepackage{natbib}
\usepackage{xcolor}
\usepackage{graphicx}
\usepackage{subfigure}
\usepackage{caption}
\usepackage{mlmath}
\usepackage{tikz}
\usetikzlibrary{shapes.geometric, arrows}
\usepackage{cleveref}       

\newtheorem{definition}{Definition}
\newtheorem{remark}{Remark}

\begin{document}

\title{Loss Spike in Training Neural Networks}

\newif\ifuniqueAffiliation
\uniqueAffiliationtrue

\ifuniqueAffiliation 
\author{ 
	{\hspace{1mm}Xiaolong Li}\thanks{
		Authors are listed in alphabetical order of last names.} \\
	School of Mathematical Sciences, Institute of Natural Sciences, MOE-LSC, Shanghai Jiao Tong University \\
	\texttt{15369855310@sjtu.edu.cn} \\
	\And
	{\hspace{1mm}Zhi-Qin John Xu}\thanks{Corresponding author: \texttt{xuzhiqin@sjtu.edu.cn}.} \\
     School of Mathematical Sciences, Institute of Natural Sciences, MOE-LSC, Shanghai Jiao Tong University \\
	Key Laboratory of Marine Intelligent Equipment and System, Ministry of Education, P.R. China \\
	\texttt{xuzhiqin@sjtu.edu.cn} \\
	\And
	{\hspace{1mm}Zhongwang Zhang} \\
	School of Mathematical Sciences, Institute of Natural Sciences, MOE-LSC, Shanghai Jiao Tong University \\
	\texttt{0123zzw666@sjtu.edu.cn}
}
\else
\usepackage{authblk}
\renewcommand\Authfont{\bfseries}
\setlength{\affilsep}{0em}
\newbox{\orcid}\sbox{\orcid}{} 
\author[1]{%
	{\usebox{\orcid}\hspace{1mm}Xiaolong Li\thanks{\texttt{15369855310@sjtu.edu.cn}}}%
}
\author[1,2]{%
	{\usebox{\orcid}\hspace{1mm}Zhi-Qin John Xu\thanks{\texttt{xuzhiqin@sjtu.edu.cn}}}%
}
\author[1]{%
	{\usebox{\orcid}\hspace{1mm}Zhongwang Zhang\thanks{\texttt{zhongwang.zhang@example.com}}}%
}
\affil[1]{School of Mathematical Sciences, Institute of Natural Sciences, MOE-LSC, Shanghai Jiao Tong University}
\affil[2]{Key Laboratory of Marine Intelligent Equipment and System, Ministry of Education, P.R. China}
\fi

\renewcommand{\shorttitle}{Loss Spike in Training Neural Networks}


\maketitle

\begin{abstract}

In this work, we investigate the mechanism underlying loss spikes observed during neural network training. When the training enters a region with a lower-loss-as-sharper (LLAS) structure, the training becomes unstable, and the loss exponentially increases once the loss landscape is too sharp, resulting in the rapid ascent of the loss spike. The training stabilizes when it finds a flat region. From a frequency perspective, we explain the rapid descent in loss as being primarily influenced by low-frequency components. We observe a deviation in the first eigendirection, which can be reasonably explained by the frequency principle, as low-frequency information is captured rapidly, leading to the rapid descent. Inspired by our analysis of loss spikes, we revisit the link between the maximum eigenvalue of the loss Hessian ($\lambda_{\mathrm{max}}$), flatness and generalization. We suggest that $\lambda_{\mathrm{max}}$ is a good measure of sharpness but not a good measure for generalization. Furthermore, we experimentally observe that loss spikes can facilitate condensation, causing input weights to evolve towards the same direction. And our experiments show that there is a correlation (similar trend) between $\lambda_{\mathrm{max}}$ and condensation. This observation may provide valuable insights for further theoretical research on the relationship between loss spikes, $\lambda_{\mathrm{max}}$, and generalization. 
\end{abstract}


\section{Introduction}
\begin{figure}[h]
	\centering
	\includegraphics[width=0.5\textwidth]{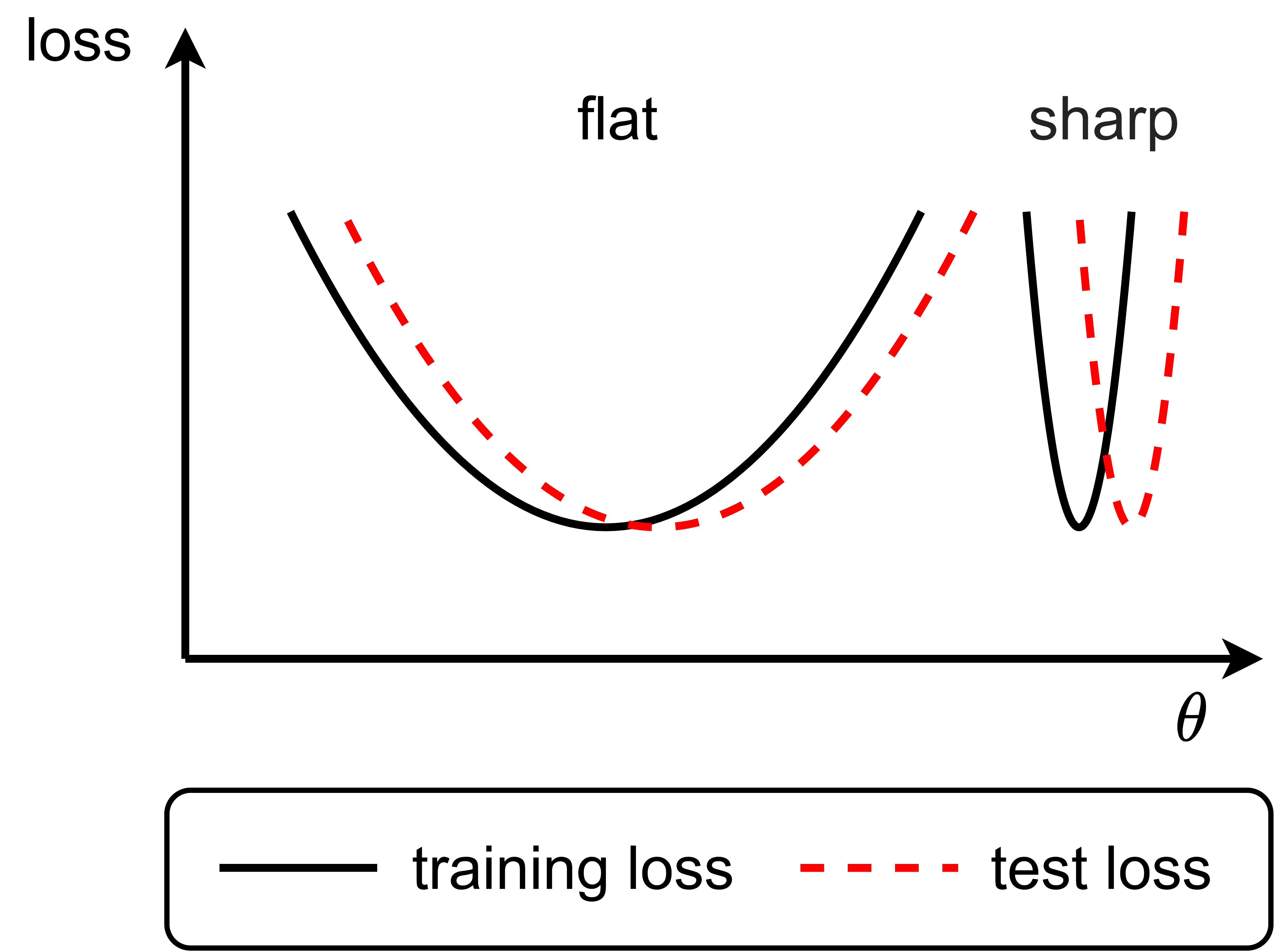}
	\caption{Schematic illustration of an ideal explanation for why flat solutions generalize well \citep{keskar2016large}.}
	\label{pic:flatness_drawio}
\end{figure}

Many experiments have observed a phenomenon, called the edge of stability (EoS) \citep{wu2018sgd, cohen2021gradient, arora2022understanding,chen2022gradient,zhu2022understanding}, that learning rate ($\eta$) and sharpness (i.e., the largest eigenvalue of Hessian) no longer behave as in traditional optimization, sharpness hovers at $2/\eta$ while the loss continues decreasing, albeit non-monotonically. Training with a larger learning rate leads to a solution with smaller $\lambda_{\rm max}$. Since $\lambda_{\rm max}$ is often used to indicate the sharpness of the loss landscape, a larger learning rate results in a flatter solution. Intuitively as shown in Fig. \ref{pic:flatness_drawio}, the flat solution is more robust to perturbation and has better generalization performance \citep{keskar2016large,hochreiter1997flat}. Therefore, training with a larger learning rate would achieve better generalization performance. In this work, we argue this intuitive analysis in Fig. \ref{pic:flatness_drawio} with  $\lambda_{\rm max}$ as the sharpness measure, which encounters difficulty in NNs through the study of loss spikes.

In a neural network training process, one may sometimes observe a phenomenon of loss spike (typical examples in Fig. \ref{pic:spike_real}), where the loss rapidly ascends and then descends to the value before the ascent. We show a special loss landscape structure underlying the loss spike, which is called a lower-loss-as-sharper (LLAS) structure. In the LLAS structure, the training is driven by descending the loss while entering an increasingly sharp region. Once the sharpness is too large, the loss would ascend exponentially fast. To explain why the loss can descend so fast, we provide a frequency perspective analysis. We find that the deviation in the ascending stage is dominated by low-frequency components. Based on the frequency principle \citep{xu2019training,xu2019frequency} that low-frequency converges faster than high-frequency, we rationalize the fast descent.  

The study of loss spike provides an important information that the deviation at the first eigendirection is dominated by low-frequency. We then further argue the link between $\lambda_{\mathrm{max}}$ flatness and generalization. In real-world datasets, low-frequency information is often dominant and well-captured by both the training data and the test data. Therefore, the training can learn low-frequency well. Since the sharpest direction, indicated by the maximum eigenvalue of the loss Hessian, relates more to the low-frequency, a solution with good generalization and a solution with bad generalization have little difference in the sharpest direction, verified by a series of experiments. Hence, $\lambda_{\mathrm{max}}$ with the intuitive explanation in Fig. \ref{pic:flatness_drawio} encounters difficulty in understanding the generalization of neural networks, such as why a larger learning rate results in better generalization for networks with EoS training.

We also find that a loss spike can facilitate condensation, that is, the input weights of different neurons in the same layer evolve towards the same, which would reduce the network's effective size. Condensation is a non-linear feature learning phenomenon in neural networks, which may be the underlying mechanism for why the loss spike improves generalization \citep{he2019control,jastrzkebski2017three}, rather than simply controlling the value of $\lambda_{\mathrm{max}}$. 

We believe that this work makes contributions in the following aspects: (1) Analyzing the loss spike phenomenon and its frequency mechanism; (2) Proposing the LLAS structure to explain the ascent stage of loss spikes; (3) Revisiting the relationship between flatness and generalization from the frequency perspective; (4) Preliminarily revealing the correlation among loss spikes, maximum eigenvalues, and the condensation phenomenon.

\section{Related works}

Previous works \citep{cohen2021gradient, wu2018sgd, xing2018walk, ahn2022understanding, lyu2022understanding, wang2022analyzing} conduct an extensive study of the EoS phenomenon under various settings. It has been observed that when the initial sharpness exceeds $2/\eta$, gradient descent ``catapults'' into a stable region and converges \citep{lewkowycz2020large}. Progressive sharpening and the edge of stability phenomenon have been analyzed under specific settings, such as normalized gradient descent \citep{arora2022understanding}. It has been shown that the third-order terms bias towards flatter minima to understand EoS \citep{damian2022self}.

The loss landscape's subquadratic structure, i.e., the maximum eigenvalue of the loss Hessian being larger when the loss is lower in a direction, has been attributed to the progressive sharpening phenomenon \citep{JML-1-247}. This work also proposes a flatness-driven motion to study the EoS stage, that is, the training would move towards a flatter minimum, such that the fixed flatness can correspond to points with lower and lower loss values due to the subquadratic property. We call this structure a lower-loss-as-flatter (LLAF) structure. The LLAF structure should expect a continuous decrease in the loss rather than a loss spike. A quadratic regression model with MSE has been used to study EoS, but in this model, the loss spike cannot occur \citep{agarwala2022second}. The loss spike has been studied from the perspective of adaptive gradient optimization algorithms \citep{ma2022qualitative}, while in this paper, the focus is on the loss landscape structure and gradient descent training.

A series of works link the generalization performance of solutions to the landscape of loss functions through the observation that flat minima tend to generalize better \citep{hochreiter1997flat, wu2017towards, ma2021linear,du2019gradient}. Algorithms that favor flat solutions are designed to improve the generalization of the model \citep{izmailov2018averaging,chaudhari2019entropy,lin2018don, zheng2021regularizing, foret2020sharpness, ding2024flat}. On the other hand, it has been shown that sharp minima can also generalize well by rescaling the parameters at a flat minimum with ReLU activation \citep{dinh2017sharp}. In this work, the relationship between flatness and generalization is studied from a new perspective, i.e., the frequency perspective, without the limitation of the activation function.

It has been identified that the linear regime and the condensed regime of parameter initialization for two-layer and three-layer wide ReLU neural networks play a crucial role in determining the network's final fitting result \citep{luo2021phase,zhou2022empirical}. The training dynamics of NNs are approximately linear and similar to a random feature model. On the contrary, in the condensed regime, active neurons are condensed at several discrete orientations. At this point, the network is equivalent to another network with a reduced width, which may explain why NNs outperform traditional algorithms \citep{breiman1995reflections,zhang2021understanding}. For the initial stage of training, A series of works \citep{zhou2021towards,chen2023phase, maennel2018gradient, pellegrini2020analytic} study the characteristics of the initial condensation for different activation functions. Du et al. \citep{du2019gradient} proved that  gradient descent achieves zero training loss in polynomial time in the linear regime of a deep overparameterized neural networks with residual connections (ResNet).

Usually, loss spikes occur when the learning rate is relatively high. This behavior is often linked to the learning dynamics of the model under different learning rates. A significant amount of work \citep{ren2024understanding} has been dedicated to exploring why neural networks trained with large learning rates for a longer time often lead to better generalization. Andriushchenko et al. \citep{andriushchenko2022sgd} finds that stochastic gradient descent (SGD) with a large learning rate can facilitate sparse solutions, attributing this effect to the noise structure inherent in SGD. It has been found \citep{li2019towards} that a large learning rate model first learns easier-to-fit patterns and is unable to memorize hard-to-fit patterns, leading to a plateau in accuracy. Once the learning rate is annealed, it is able to fit these patterns, explaining the sudden spike in both train and test accuracy. In our work, we find that for the noise-free full-batch gradient descent algorithm, the loss spike can also facilitate the condensation phenomenon, implying that the noise structure is not the intrinsic cause of condensation.

The frequency principle is examined in extensive datasets
and deep neural network models \citep{xu2019training, xu2021deep,rahaman2018spectral}. Subsequent theoretical studies show that the frequency principle holds in the general setting with infinite samples \citep{luo2019theory}. An overview for frequency principle is referred to Xu et al. \citep{xu2022overview}.  Based on the theoretical understanding, the frequency principle inspires the design of deep neural networks to learn a function with high-frequency fast \citep{liu2020multi,jagtap2020adaptive,biland2019frequency}.



\section{Loss spike}
\subsection{Preliminary: Linear stability in training quadratic model}
We consider a simple quadratic model with the loss $R(\theta) = \lambda\theta^2/2$ trained by gradient descent with learning rate $\eta$, $ \theta(t+1) = \theta(t)-\eta \cdot d R(\theta)/ d \theta$.
To ensure the linear stability of the training, it is required that $|\theta(t+1)|<|\theta(t)|$, which implies $|1-\lambda \eta|<1$. Otherwise, the training will diverge. Note that $\lambda$ is the Hessian of $R(\theta)$. Similarly, to ensure the linear stability of training a neural network, it requires that the maximum eigenvalue of the loss Hessian is smaller than $2/\eta$, i.e., 2 over the learning rate. Therefore, the maximum eigenvalue of the loss Hessian is often used as the measure of the sharpness of the loss landscape.
In this section, we study the phenomenon of loss spike, where the loss would suddenly increase and decrease rapidly. For example, as shown in Fig. \ref{pic:spike_real} (a, d, g, h), we train a tanh fully-connected neural network (FNN) with 20 hidden neurons for a one-dimensional fitting problem, a ReLU convolutional neural network (CNN) for the CIFAR10-1k (a subset of the well-known CIFAR-10 dataset, which includes 1,000 images from 10 different classes) classification problem with MSE, and two VGG-11 \citep{simonyan2014very} models with different learning rates for the CIFAR-10 dataset with datasize 1024. These models experience loss spikes. The red curves, i.e., the $\lambda_{\rm max}$ value, show that the loss spikes occur at the EoS stage. 

\subsection{Typical loss spike experiments}
\begin{figure}[h!]
	\centering
         \subfigure[FNN, full loss]{\includegraphics[width=0.32\textwidth]{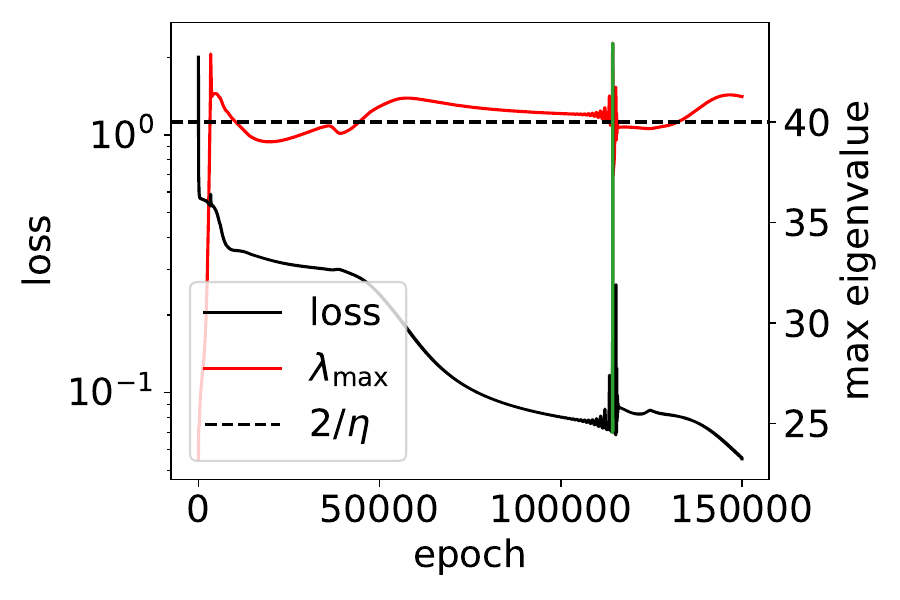}}	
         \subfigure[FNN, loss]{\includegraphics[width=0.32\textwidth]{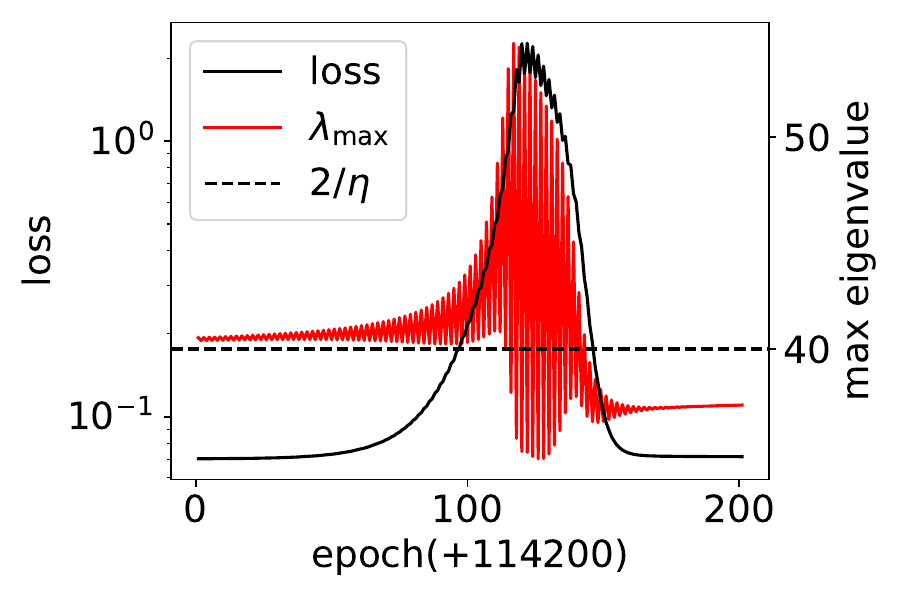}}
        \subfigure[FNN, trajectory]{\includegraphics[width=0.32\textwidth]{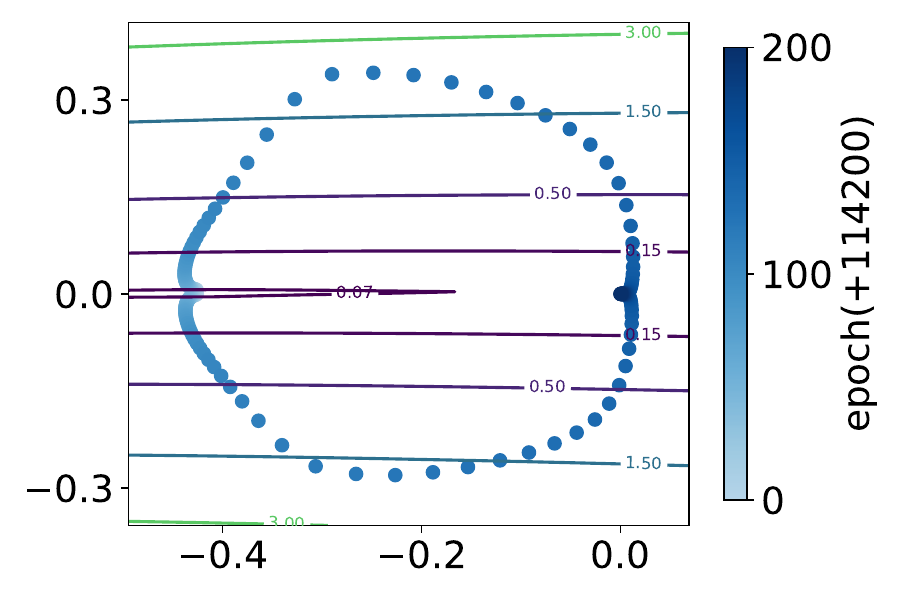}}  
         \subfigure[CNN, full loss]{\includegraphics[width=0.32\textwidth]{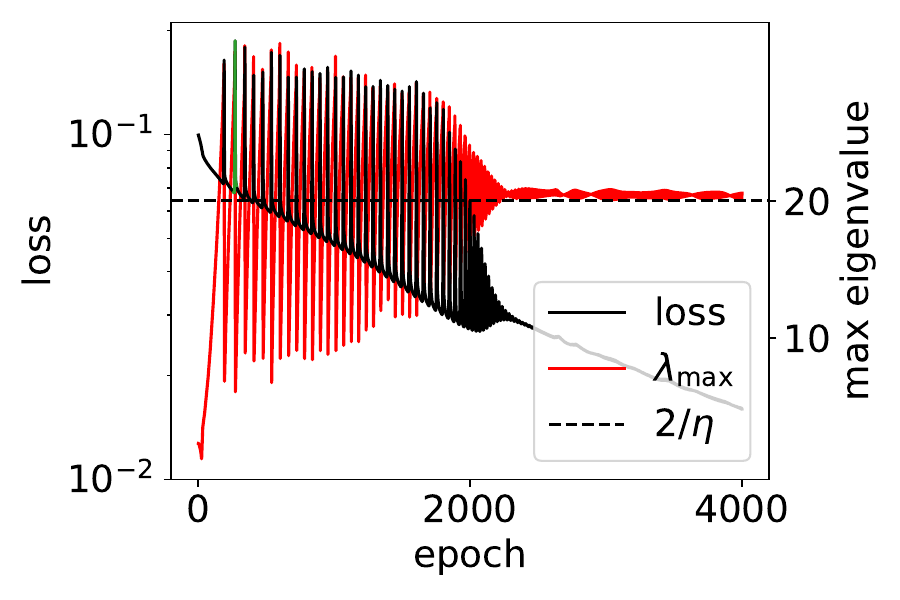}}
	\subfigure[CNN, loss]{\includegraphics[width=0.32\textwidth]{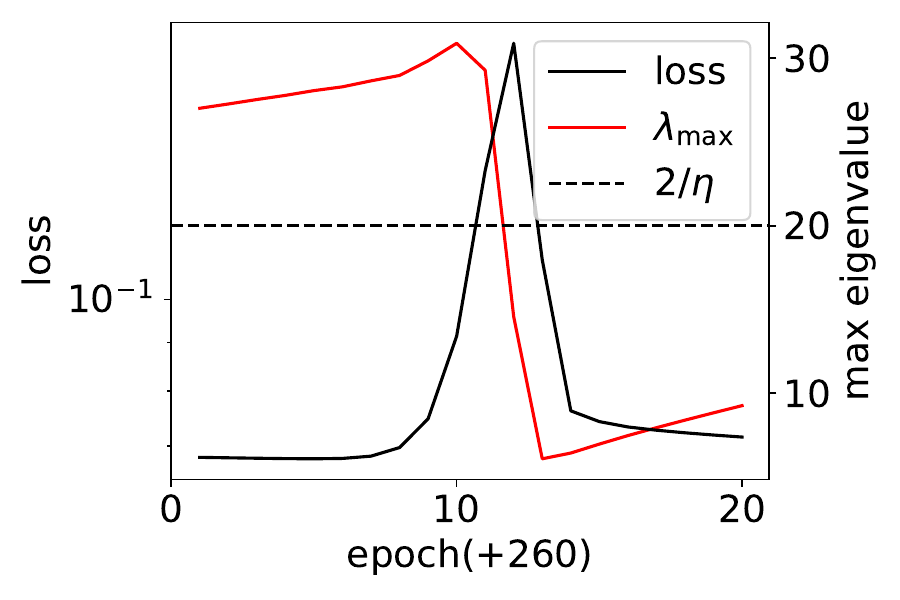}}
	\subfigure[CNN, trajectory]{\includegraphics[width=0.32\textwidth]{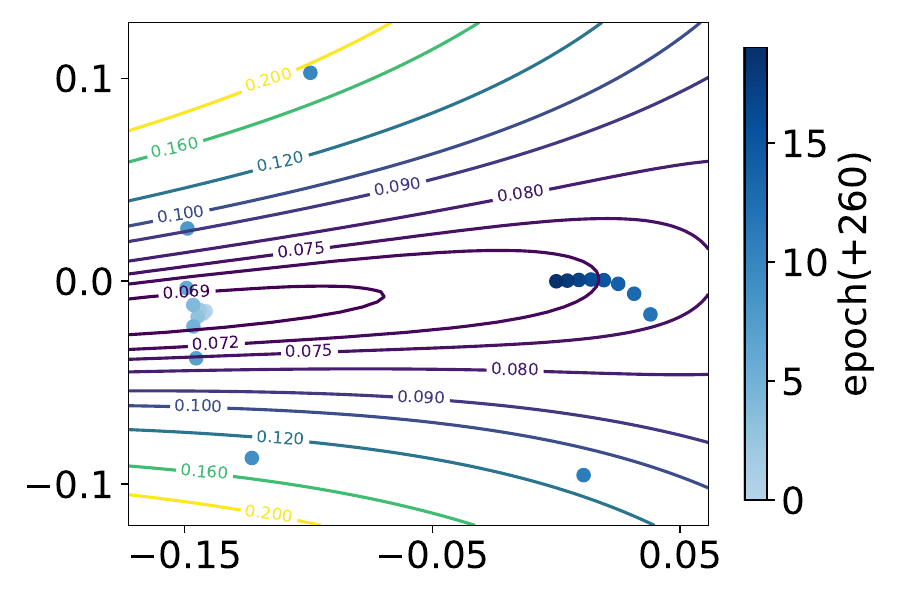}}
        \centering
         \subfigure[VGG, lr=0.2]{\includegraphics[width=0.31\textwidth]{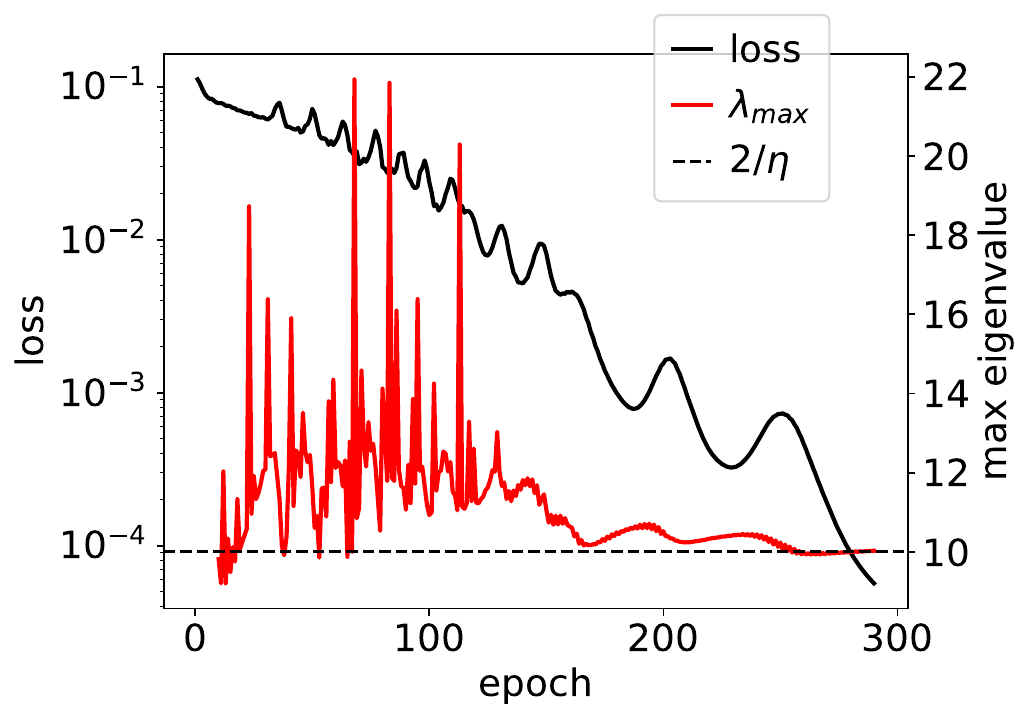}}	
         \subfigure[VGG, lr=0.3]{\includegraphics[width=0.31\textwidth]{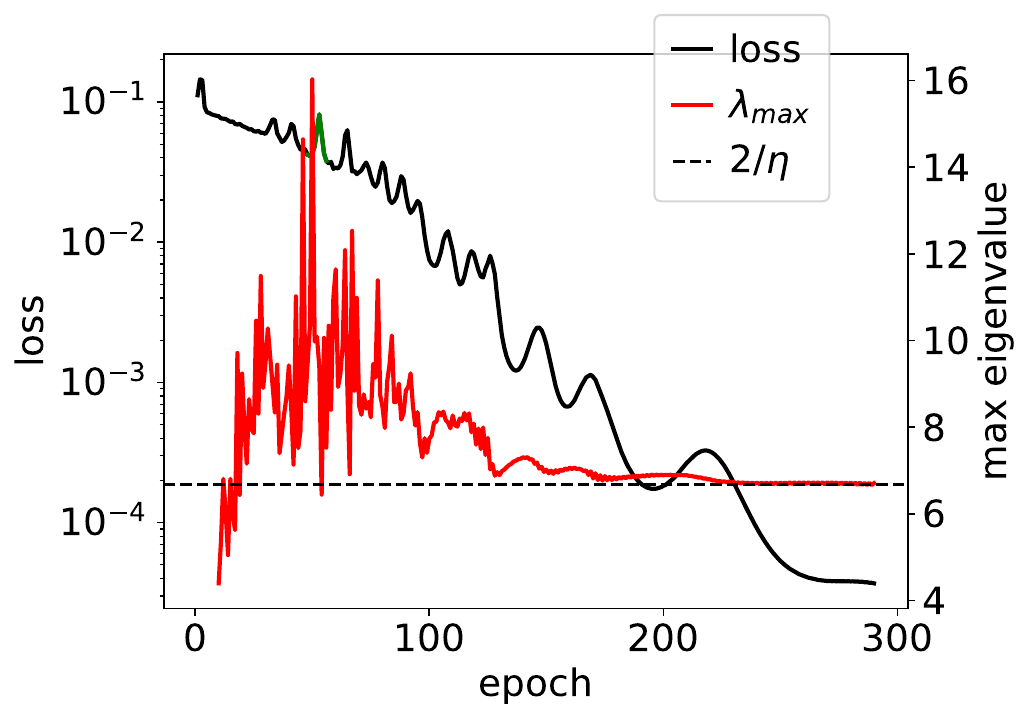}}
        \subfigure[VGG, trajectory]{\includegraphics[width=0.3\textwidth]{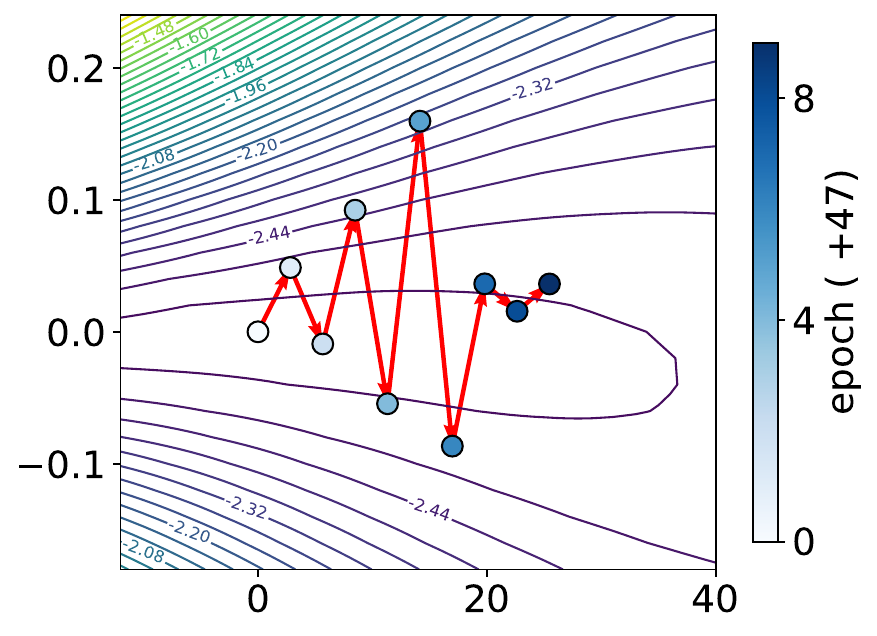}}

    \caption{ (a, d, g,  h) The loss value (black) and $\lambda_{\rm max}$ (red) vs. training epoch. (b, e) The loss value and $\lambda_{\rm max}$ of a specific epoch interval, which is marked green in (a, d), respectively. (c, f, i) The loss surface and the trajectory of the model parameters along the first two PCA directions. (a, b, c) Two-layer tanh NN with width 20. The sum of the explained variance ratios of the first two PCA directions is 0.9895. (d, e, f) Two-layer ReLU CNN with Max Pooling. The sum of the explained variance ratios of the first two PCA directions is 0.9882. (g, h, i) VGG-11 \citep{simonyan2014very} with different learning rates. The sum of the explained variance ratios of the first two PCA directions is 0.9999.}
    \label{pic:spike_real}
\end{figure}

To observe the loss spike clearly, we zoom in on the training epochs around the spike, shown in Fig. \ref{pic:spike_real} (b, e). The selected epochs are marked green in Fig. \ref{pic:spike_real} (a, d). When the maximum eigenvalue of Hessian $\lambda_{\mathrm{max}}$ (red) exceeds $2/\eta$ (black dashed line), the loss increases, and when $\lambda_{\mathrm{max}}<2/\eta$, the loss decreases, which are consistent with the linear stability analysis.




We then study the parameter space for more detailed characterization. Given $t$ training epochs, and let $\vtheta_i$ denote model parameters at epoch $i$, we apply PCA to the matrix $M = [\vtheta_{1}-\vtheta_{t}, \cdots, \vtheta_{t}-\vtheta_{t}]$, and then select the first two eigendirections $\ve_1$, $\ve_2$. The two-dimensional loss surface based on $\ve_1$ and $\ve_2$ can be calculated by $R_S(\vtheta_{t}+\alpha \ve_1 + \beta \ve_2)$, where $\alpha$, $\beta$ are the step sizes, and $R_S$ is the loss function under the dataset $S$. The trajectory point of parameter $\vtheta_i$ can be calculated by the projection of $\vtheta_{i}-\vtheta_{t}$ in the PCA directions, i.e., $( \langle \vtheta_{i}-\vtheta_{t},\ve_1\rangle, \langle \vtheta_{i}-\vtheta_{t},\ve_2 \rangle)$. Parameter trajectories (blue dots) and loss surfaces along PCA directions are shown in Fig. \ref{pic:spike_real} (c, f, i). In three distinct examples, they exhibit similar behaviors. At the beginning of the ascent stage of the spike, the parameter is at a small-loss region, where the opening of the contour lines is towards the left, indicating a leftward component of descent direction. In the left region, the contour lines are denser, implying a sharper loss surface. Once $\lambda_{\rm max}>2/\eta$, the parameters become unstable, and the loss value increases exponentially. In the high-loss region, the opening of the contour shifts to the right, indicating a rightward component of the descent direction, resulting in a sparser contour, i.e., a flatter loss surface. After several steps,  when $\lambda_{\rm max}<2/\eta$, the training returns to the stable stage.

\subsection{Lower-loss-as-sharper (LLAS) structure}
\begin{figure}[h]
	\centering
        \subfigure[LLAS example]{\includegraphics[width=0.45\textwidth]{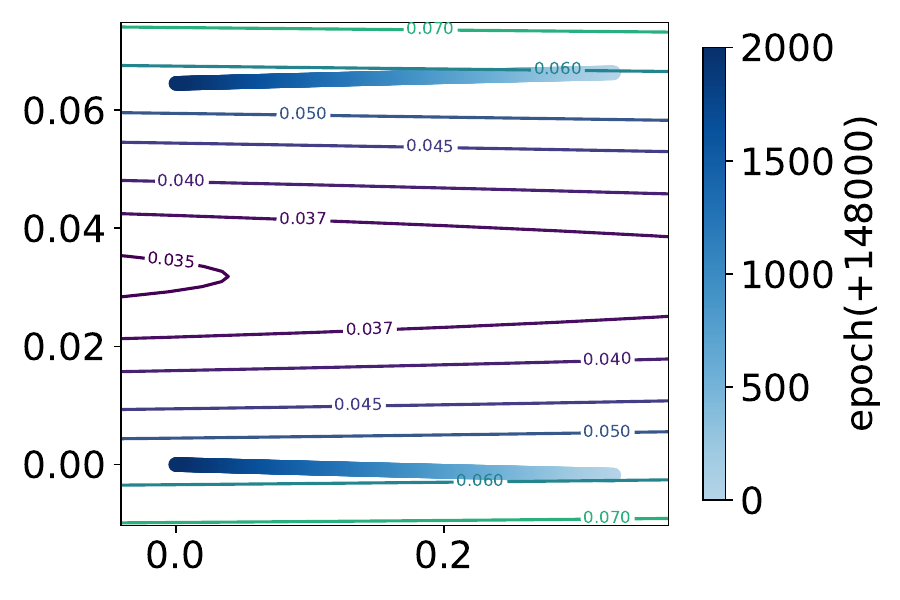}
        \label{fig: LLAS}}
	\subfigure[schematic illustration]                         {\includegraphics[width=0.45\textwidth]      {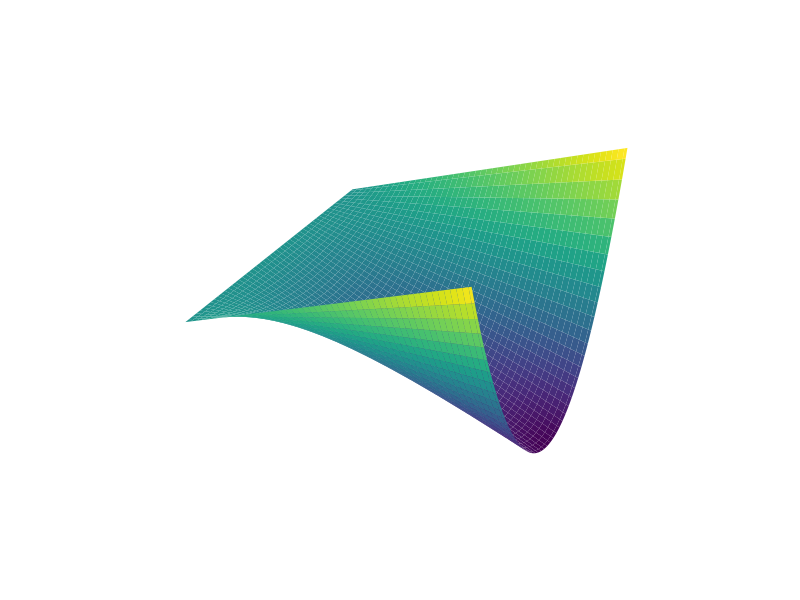}
        \label{fig: toy_a}}
	\subfigure[toy model, loss]{\includegraphics[width=0.45\textwidth]{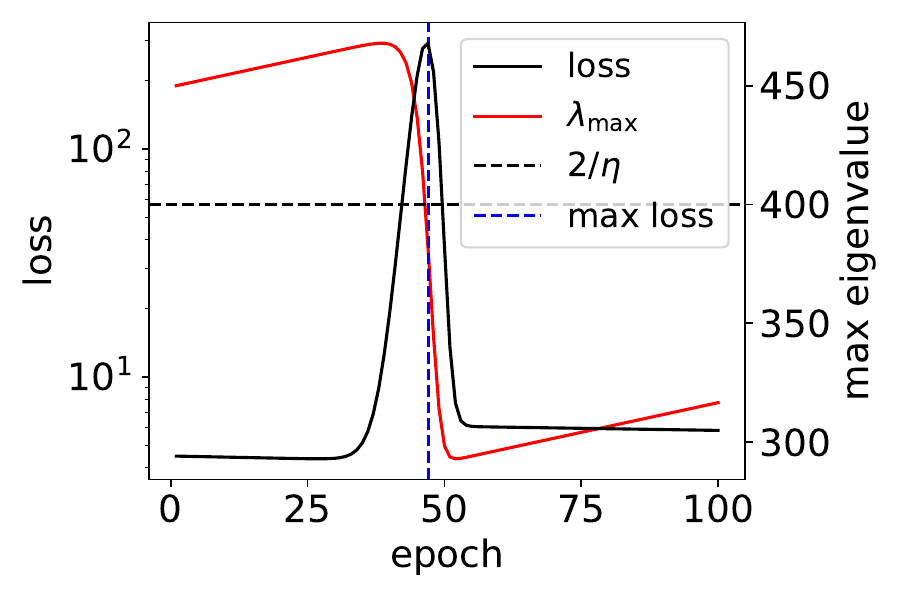}
    \label{fig: toy_loss}}
        \subfigure[toy model, trajectory]{\includegraphics[width=0.45\textwidth]{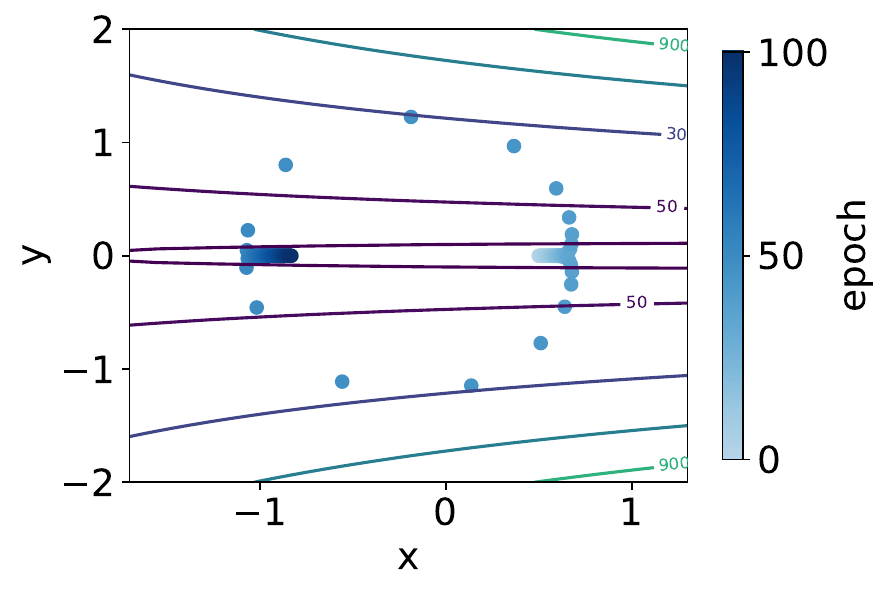}
        \label{fig: toy_trajectory}}
    \caption{ (a) The loss surface and the trajectory of the model parameters along the first two PCA directions in the EoS stage. (b) Schematic illustration of LLAS structure in 3D. (c) The loss value and the maximum eigenvalue of the Hessian matrix of a loss spike process of the toy model. (d) The loss surface and the GD trajectory of the two-dimensional parameters of the toy model.}  \label{pic:spike_toy}
\end{figure}

The above experiments reveal a common structure that causes a loss spike, namely, the $\lambda_{\mathrm{max}}$ sharpness increases in the direction of decreasing loss. We call this structure lower-loss-as-sharper (LLAS) structure. The LLAS structure, which is common in the EoS stage as shown in Fig. \ref{fig: LLAS}, differs from the LLAF (lower-loss-as-flatter) structure studied in \citep{JML-1-247}. The following quadratic model is a simple example of the LLAS structure.

\begin{equation}
    f(x,y)=(50 x+200)y^2-x+5,
    \label{equ:toy_model}
\end{equation}
where $(x,y) \in (-4, + \infty) \times \mathbb{R}$.

For any constant $C$, $y=0$ is the minimum point of $f(C, y)$. The larger $x$ is, the lower the loss value is, and the sharper the loss landscape in the $y$-direction is. The intuitive explanation for the above phenomenon is that as $x$ increases, $f(x, 0)$ decreases, which means that $f(x, 0)$ has a smaller value at the sharp region, i.e., the LLAS structure, which makes the opening of the contour lines towards different directions at different loss levels.

This model can be considered as being obtained by constraining the parameters of the linear neural network.

The toy example of the LLAS structure is shown in Fig. \ref{pic:spike_toy} (b). As shown in Fig. \ref{pic:spike_toy} (c, d), the loss curve and the trajectory of parameters are similar to the realistic example above, where the parameters move toward the sharp direction at the beginning of the loss spike, and then move toward the flat direction.

For this example, we can exactly compute the derivative of Eq. (\ref{equ:toy_model})  as follows:
\begin{equation*}
    \frac{\partial f(x,y)}{\partial x}=50 y^2-1. 
\end{equation*}
\begin{equation*}
    \frac{\partial f(x,y)}{\partial y}=(100 x+400)y. 
\end{equation*}

Thus we have
\begin{equation*}
    \frac{\partial f(x,y)}{\partial x} \begin{cases}>0 & \text { if } f(x,y) >9 \\ =0 & \text { if  } f(x,y) =9 \\ <0 & \text { if  } f(x,y) <9 \end{cases}, 
\end{equation*}
which indicates that the toy model has a negative gradient component in the $x$ direction when the parameters are in the small-loss region ($f(x,y)<9$), while a positive gradient component in the $x$ direction when the parameters are in the high-loss region ($f(x,y)>9$) (more details in Appendix \ref{app:LLAS_further}).

Although the LLAS structure can explain the mechanism of the ascent stage based on the toy model, it can not explain the reason for the rapid descent of the loss in the descent phase of the loss spike, which takes much fewer steps than the training from the same level loss at the initialization. Moreover, due to the high dimensionality of the parameter space, the parameter trajectory does not always align with the first eigendirection, otherwise, as shown in the toy model, the loss would not decrease continuously. In the following, we take a step toward understanding the rapid decrease from the frequency perspective.

\subsection{Frequency perspective for understanding descent stage}

In this subsection, we study the mechanism of the rapid loss descent during the descent stage in a loss spike from the perspective of frequency. The “frequency” means response frequency which is the frequency of a general Input-Output mapping \citep{xu2019frequency}.




Our analysis is based on the commonly observed phenomenon known as the frequency principle \citep{xu2019training,xu2019frequency,zhang2021linear,luo2019theory,rahaman2018spectral,basri2019convergence}, which states that deep NNs often fit target functions from low to high frequencies during the training. Compared to the peak point of the loss spike with the point with the same loss value at the initial training, the descent during the spike should eliminate more low-frequency component with a fast speed while the descent from the initial model should eliminate more high-frequency component with a slow speed. To verify this conjecture, we study the frequency distribution of the converged part during the descent stage.


\begin{figure}[h!]
	\centering
        \subfigure[low-frequency threshold]
	{\includegraphics[width=0.4\textwidth]{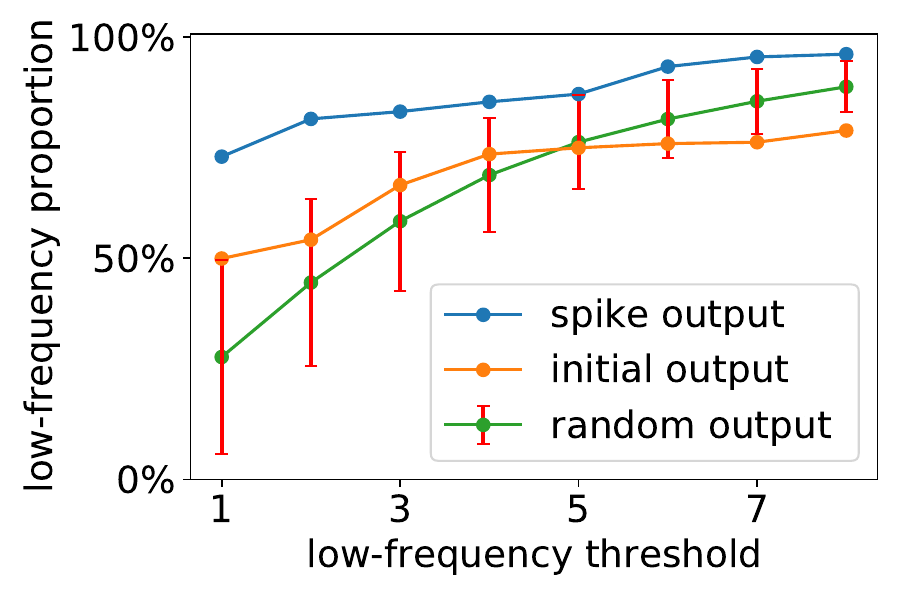}}
	\subfigure[low-frequency and loss]     {\includegraphics[width=0.4\textwidth]      {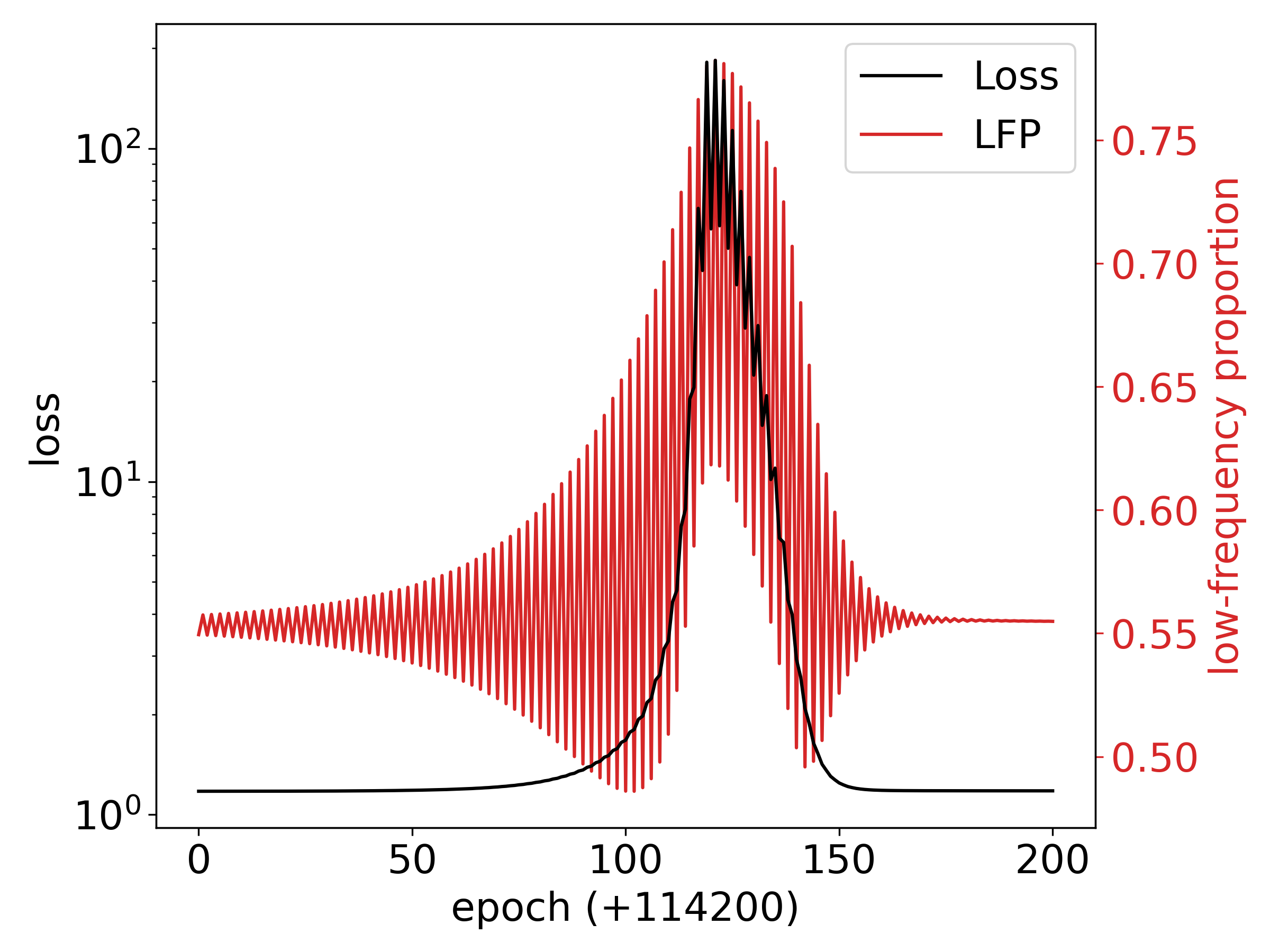}}
  \caption{(a) Low-frequency proportion for different low-frequency thresholds. The NN we used is a two-layer tanh NN with width 20. For the random output difference, we calculate the mean value and the error bar with 100 random samples. (b) Train loss and low-frequency proportion for low-frequency threshold = 2 for different epoch.}
  \label{pic:errorbar}
\end{figure} 

The parameter at the peak of the loss spike is denoted as $\vtheta_{\mathrm{max}}$, and the parameter at the initial point which has the similar loss to the loss of $\vtheta_{\mathrm{max}}$
is denoted as  $\vtheta_{\mathrm{ini,m}}$, the parameter at the end of the loss spike\footnote{This point is selected roughly during the slow descent because, experimentally, it does not affect the result.} is denoted as $\vtheta_{\mathrm{end}}$. We then study the frequency distribution of spike output difference $f_{\rm peak,diff}:=f_{\vtheta_{\mathrm{max}}}-f_{\vtheta_{\mathrm{end}}}$ and initial output difference $f_{\rm ini,diff}:=f_{\vtheta_{\mathrm{ini,m}}}-f_{\vtheta_{\mathrm{end}}}$. 

For comparison, we also randomly construct a parameter (ensuring that the magnitude of the random perturbation is consistent with the parameter magnitude at the loss spike), and then we use it as a control experiment to compare the frequency of differences in output:
$$
\vtheta_{\mathrm{rnd}} := \vtheta_{\mathrm{end}} + \left(\frac{\norm{\vtheta_{\mathrm{end}} - \vtheta_{\mathrm{max}}}_2}{\norm{\varepsilon}_{2}}\right) \varepsilon,
$$
where \(\varepsilon \sim N(0, I)\) is a random variable.

We then study the frequency distribution of the random output difference 
$
f_{\rm rnd,diff} := f_{\vtheta_{\mathrm{rnd}}} - f_{\vtheta_{\mathrm{end}}}.
$
This comparison is made to demonstrate that the low-frequency proportion of the parameters at points where a loss spike occurs is significantly higher than in situations where no loss spike is observed, even exceeding the range of a random perturbation.

We characterize the frequency distribution by taking different low-frequency thresholds to study low-frequency proportion. 

\begin{definition}
    For a low-frequency threshold $K$, a low-frequency proportion (LFP) is defined as follows to characterize the power proportion of the low-frequency component over the whole spectrum,
    \begin{equation}
        \mathrm{LFP}(K)=\frac{\sum_{k \leq K} \norm{\hat{f}_{\vtheta}(k)}^2}{\sum_{k} \norm{\hat{f}_{\vtheta}(k)}^2},
        \label{equ:lfp}
    \end{equation}
    where $\hat{f}_{\vtheta}$ indicates the Fourier transform of function $f_{\vtheta}$.

\end{definition}


As shown in Fig. \ref{pic:errorbar} (a), the low-frequency proportion of the spike output difference is significantly larger than the low-frequency proportion of the initial output difference and the random output difference, where we take 100 samples of random variable $\varepsilon$ for the mean value and the error bar for each low-frequency threshold. Fig. \ref{pic:errorbar} (b) shows the training loss and the low-frequency proportion over epochs, where the low-frequency threshold is 2. This illustrates that the low-frequency proportion of the output increases dramatically with the occurrence of the loss spike. Fig. \ref{pic:errorbar} verifies that the higher low-frequency proportion of the spike output difference is the key reason for the rapid drop in the loss value during the descent stage, as suggested by the frequency principle.

\section{Revisit the flatness-generalization picture}

Motivated by the loss spike analysis from the frequency perspective, we further revisit the common flatness-generalization picture. A series of previous works \citep{hochreiter1997flat,li2017visualizing} attempt to link the flatness of the loss landscape with generalization, so as to characterize the model through flatness conveniently. A classic empirical illustration is shown in Fig. \ref{pic:flatness_drawio}, which vividly expresses the reason why flat solutions tend to have better generalization. Usually, the training loss landscape and the test landscape do not exactly coincide due to sampling noise. A flat solution would be robust to the perturbation while a sharp solution would not. For such a one-dimensional case, this analysis is valid, but the loss landscape of a NN case is very high-dimensional, and such simple visualization or explanation is yet to be validated.

The first eigendirection of the loss Hessian, i.e., the eigendirection corresponding to the maximum eigenvalue, is the sharpest direction. Based on the flatness-generalization picture, it is natural to use the maximum eigenvalue as the measure for the flatness, which can also indicate generalization. However, this naive analysis is not always correct for neural networks. 

\subsection{Frequency perspective}

The maximum eigenvalue of the loss Hessian can indicate the linear stability of the training process, and is often used as a measure for flatness/sharpness of the loss landscape, where a larger maximum eigenvalue indicates a sharper landscape. As shown by linear stability analysis, once the maximum eigenvalue exceeds $2/\eta$ ($\eta$ is the learning rate), the training would oscillate and diverge along the first eigendirection. Meanwhile, as the parameters move away from the minimum point along this first eigendirection, the loss spike arises mainly due to the large low-frequency difference, as shown in Fig. \ref{pic:errorbar}. Therefore, the deviation in the first eigendirection of the loss Hessian primarily leads to the deviation of low-frequency components.

In order to examine the above analysis, we first obtain the model parameter $\vtheta_{\rm train}$ with poor generalization by training the model initialized in the linear regime \citep{luo2021phase}, and then further train the model parameter $\vtheta_{\rm train}$ on the test dataset with a small learning rate to obtain the model parameter $\vtheta_{\rm test}$.

Let $\vH$ be the Hessian matrix of the loss function at the training parameters $\vtheta_{\rm train}$. $\lambda_i$ is the $i$-th eigenvalue of the Hessian matrix $\vH$ with $\lambda_1 \geq \lambda_2 \geq \cdots \geq \lambda_N$, and $\vnu_i$ is the corresponding eigendirection. We study the impact of each eigendirection on the test loss by eliminating the difference between $\vtheta_{\rm train}$ and $\vtheta_{\rm test}$ in the $i$-th eigendirection $\vnu_i$, where $i$ is the index of eigenvalues.

\begin{figure}[h]
	\centering
        \subfigure[FNN]{\includegraphics[height=0.2\textheight]{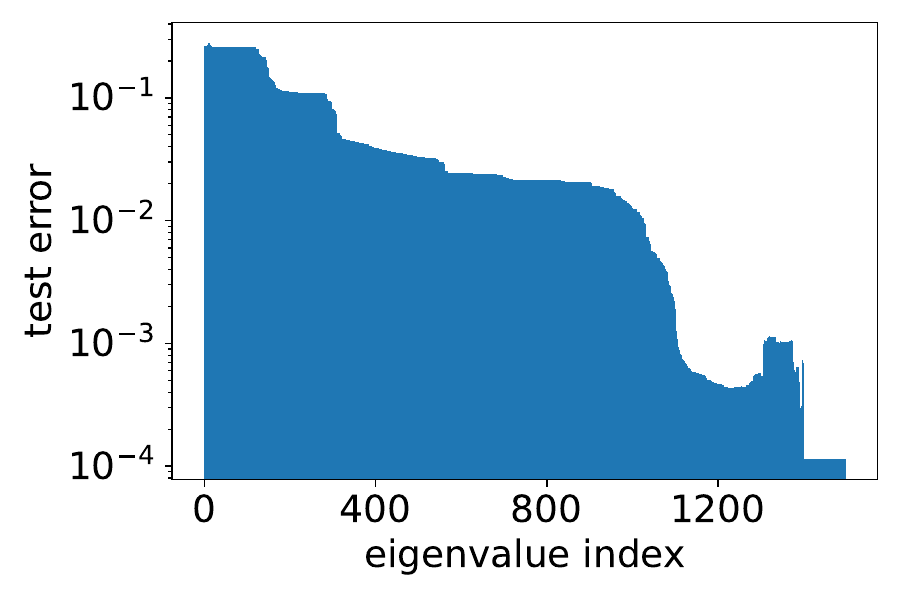}}
	\subfigure[output difference]{\includegraphics[height=0.23\textheight]{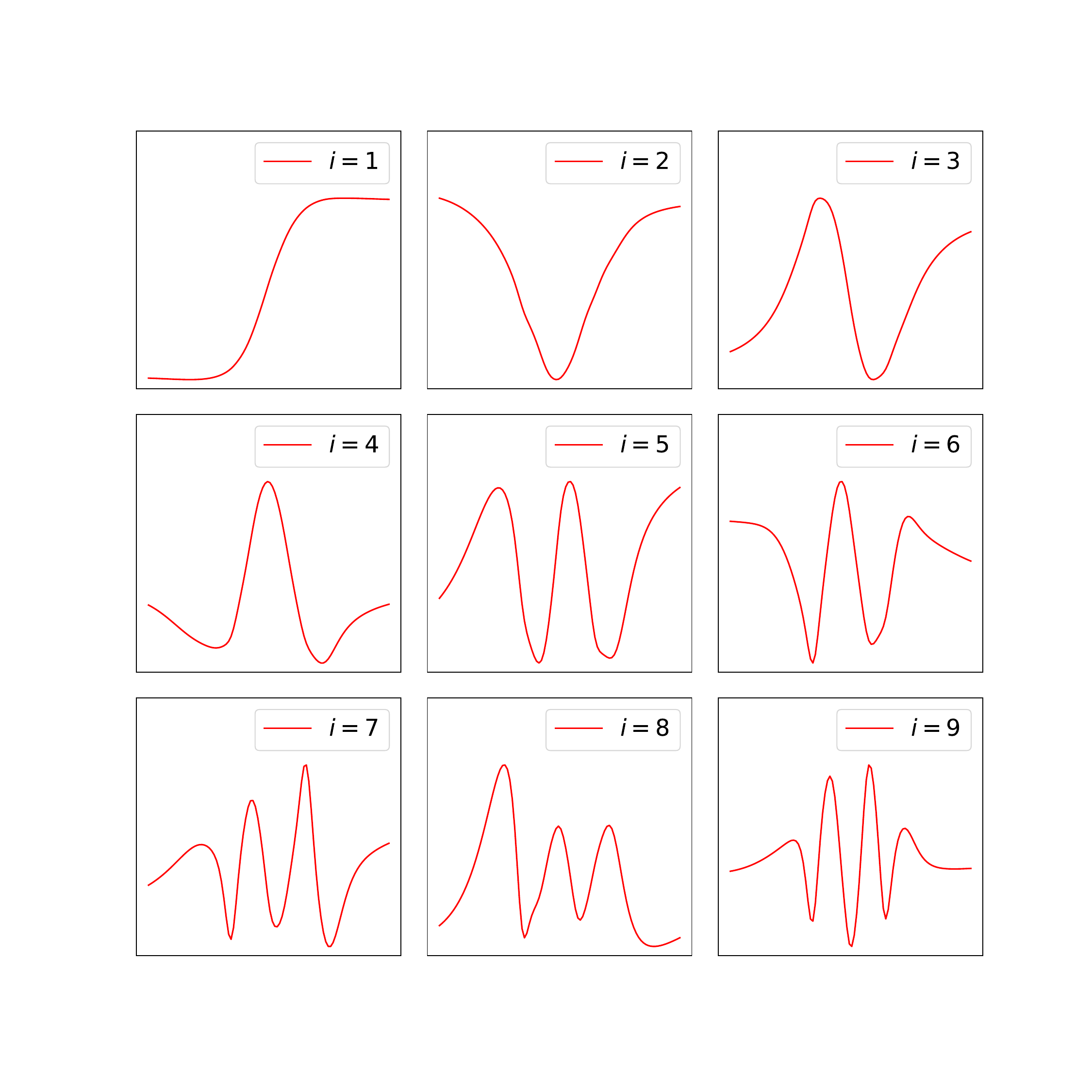}}
	
    \caption{Two-layer tanh FNN with a width of 500. (a) The variation of the test loss with the eigenvalue index $i$ when eliminating the difference between $\vtheta_{\rm train}$ and $\vtheta_{\rm test}$ in the first $i$ eigendirections. (b) The output difference before and after moving $\vtheta_{\theta_{\rm train}}$ in the first nine eigendirections of its Hessian matrix. Each subset corresponds to the case of one eigendirection.} \label{pic:frequency}
\end{figure}

As shown in Fig. \ref{pic:frequency} (a), we study the change of the test loss $L(i)$ with the eigenvalue index $i$ as follows to study the effect of eigenvectors on generalization, 
\begin{equation*}
L(i)=R_{S_{\rm test}}\left(\vtheta_{\rm train}+\sum_{j=1}^{i} \langle\vtheta_{\rm test}-\vtheta_{\rm train} , \vnu_{j}\rangle \vnu_{j}\right),
\end{equation*}
where $S_{\rm test}$ is the test dataset. The movement of parameters on the eigenvectors corresponding to large eigenvalues has a weak impact on the test loss, while the movement of parameters on the eigenvectors corresponding to small eigenvalues has a significant impact on the test loss.

A reasonable explanation from the perspective of frequency is as follows. In common datasets, low-frequency components often dominate over high-frequency ones. For noisy sampling, the dominant low-frequency is shared by both the training and the test data. When the parameters move along the eigendirections corresponding to the large eigenvalues, the network output often changes at low-frequency, which is already captured by both $\vtheta_{\rm train}$ and $\vtheta_{\rm test}$. Therefore, the improvement of model generalization often requires certain high-frequency changes. As shown in Fig. \ref{pic:frequency} (b), we move the corresponding $\vtheta_{\rm train}$ along the first nine eigendirections, and show the difference between the network outputs before and after the movement, i.e., $f_{\vtheta_{\rm train}+ \vnu_i/\sqrt{\lambda_{i}}}-f_{\vtheta_{\rm train}}$, where the $1/\sqrt{\lambda_{i}}$ item is to make the loss of the network moved in different eigendirections approximately the same. From the difference between the outputs before and after the movement, it can be seen that when the parameters move along the eigendirection corresponding to the larger eigenvalue, the change of the model output is often less oscillated, i.e., dominated by the lower-frequency.
Since the low-frequency is captured by both $\vtheta_{\rm train}$ and $\vtheta_{\rm test}$, they should be close in the eigendirections corresponding to  large eigenvalues, which is verified in the following subsection.

\subsection{Difference on each eigendirection}

\begin{figure}[h]
	\centering
        \subfigure[FNN]{\includegraphics[width=0.24\textwidth]{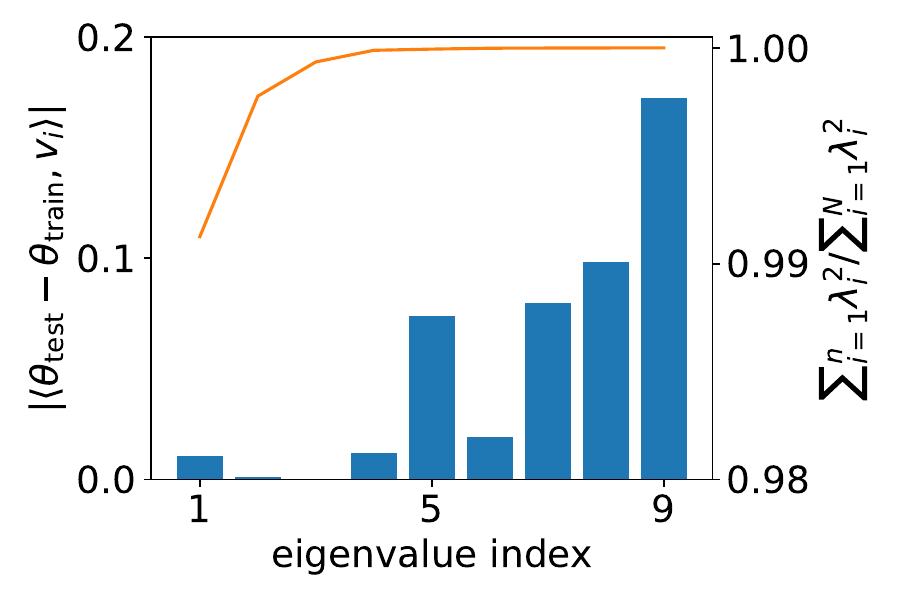}}
	\subfigure[Two-layer CNN]{\includegraphics[width=0.24\textwidth]{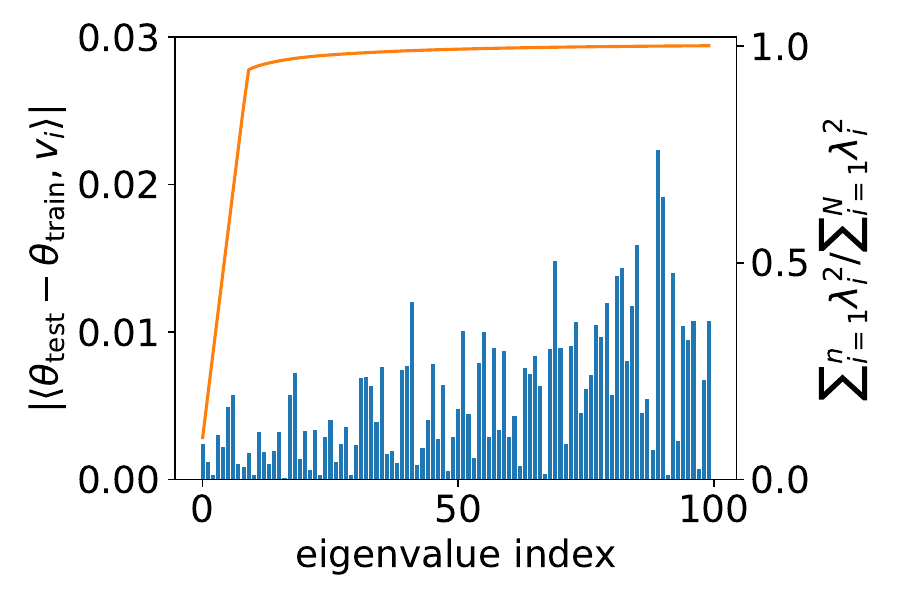}}
        \subfigure[Three-layer CNN]{\includegraphics[width=0.24\textwidth]{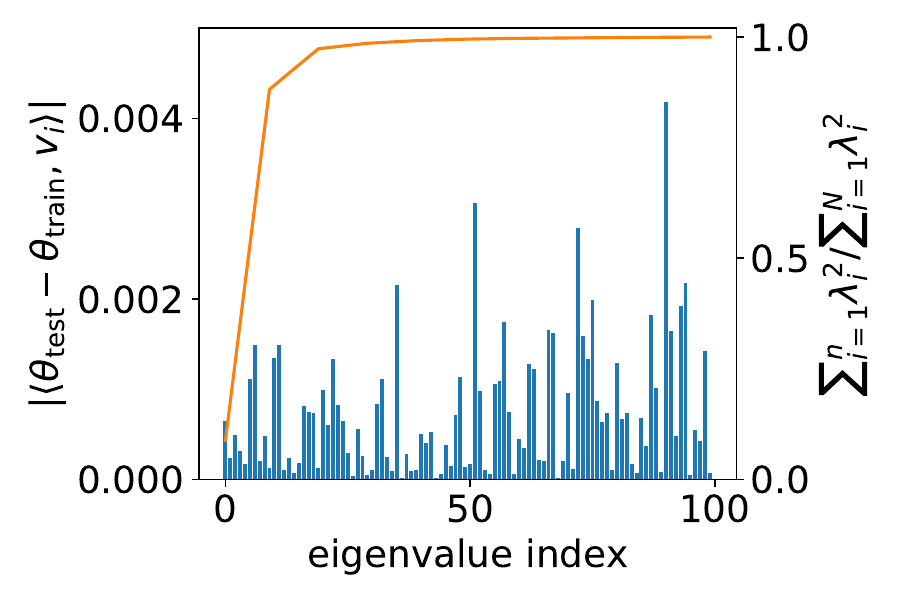}}
        \subfigure[Different batch sizes]{\includegraphics[width=0.24\textwidth]{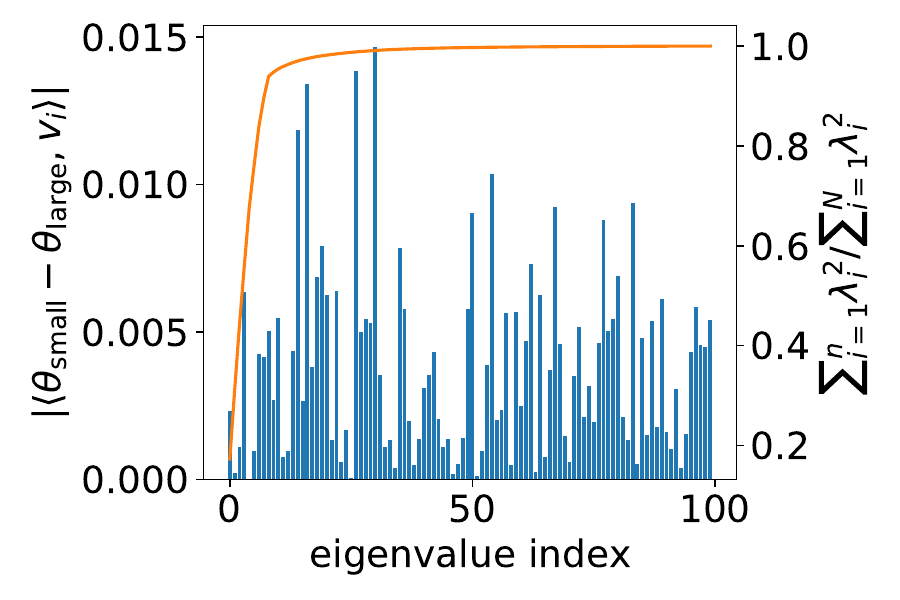}}
	
    \caption{ Blue bar: (a, b, c) show the projection values of in each eigendirection of $H(\vtheta_{\rm train})$ for $\vtheta_{\rm test}-\vtheta_{\rm train}$,  and (d) for $\vtheta_{\mathrm{large}}-\vtheta_{\mathrm{small}}$. Orange line: the sum of the first $n$ eigenvalues over all eigenvalues. (a) Two-layer tanh FNN for the one-dimensional fitting problem. (b)  Two-layer ReLU CNN with Max Pooling for the CIFAR10 classification problem. (c)  Three-layer ReLU CNN with Max Pooling for the CIFAR10 classification problem. (d)  Five-layer ReLU CNN with Max Pooling for the CIFAR10 classification problem. }
    \label{pic:projection}
\end{figure}

We then examine the projection  of $\vtheta_{\rm test}-\vtheta_{\rm train}$ in each eigendirection of $H(\vtheta_{\rm train})$.
As shown in Fig. \ref{pic:projection}, we show the projection of $\vtheta_{\rm test}-\vtheta_{\rm train}$ on each eigenvector $\vnu_i$ (blue bar) for the FNN on function fitting problem and the CNNs on CIFAR10 classification problem. Due to the high complexity of calculating the eigenvectors of the large-size Hessian matrix, we use the Lanczos method \citep{cullum2002lanczos} to numerically compute the first $N$ eigenvalues and their corresponding eigenvectors. For $s < N$, we use ${\sum}_{i=1}^{s}\lambda_i^2/{\sum}_{i=1}^{N}\lambda_i^2$ to represent the explained variance ratio, i.e., to measure how much flatness information the first $n$ eigendirections (orange line) can explain. For different network structures and model tasks, the projection value of $\vtheta_{\rm test}-\vtheta_{\rm train}$ on the eigenvector $\vnu_i$ has a positive correlation with the eigenvalue index $i$, which confirms that $\vtheta_{\rm train}$ and $\vtheta_{\rm test}$ have little difference on low-frequency part. It has been verified through a series of studies \citep{jastrzkebski2017three} that models with different batch sizes exhibit varying generalization abilities. Note that in Fig. \ref{pic:projection} (d), the two minima, $\vtheta_{\mathrm{small}}$ and $\vtheta_{\mathrm{large}}$, found by small and large batch sizes, respectively, have little difference in eigendirections corresponding to the largest $s$ eigenvalues which are enough that ${\sum}_{i=1}^{s}\lambda_i^2/{\sum}_{i=1}^{N}\lambda_i^2$ is close to 1. This indicates that the differences in the eigendirections corresponding to the largest eigenvalues are not significant, which is consistent with the conclusions drawn from Fig. \ref{pic:projection} (a, b, c).

\subsection{Implications}

\begin{figure}[h!]
	\centering
	\includegraphics[width=0.5\textwidth]{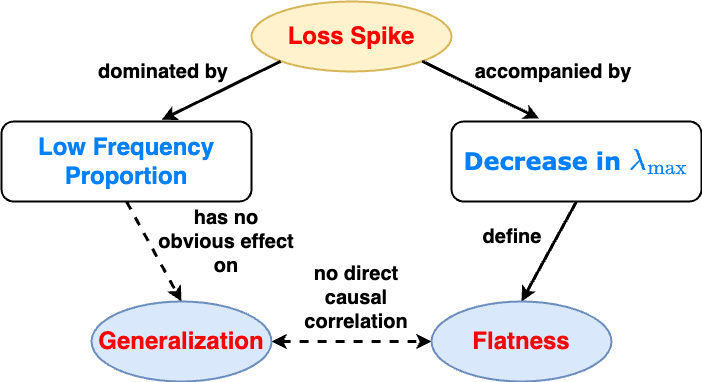}
  \caption{Flatness-Generalization Relationship}
  \label{pic:revisit}
\end{figure}

As shown in Fig. \ref{pic:revisit}, we revisit the link between $\lambda_{\mathrm{max}}$, flatness and generalization. First, we experimentally demonstrate that the difference in model output during the loss spike is mainly dominated by low-frequency components (Section 4.1), and these low-frequency components often have no significant impact on the model's generalization ability (Section 4.2). Additionally, the loss spike is often accompanied by a decrease in lambda max (as concluded in Section 3), indicating an improvement in flatness based on the maximum eigenvalue. Therefore, our core conclusion is that flatness based on the maximum eigenvalue cannot directly characterize the model's generalization ability.

The above analysis suggests the following implications: i) The maximum eigenvalue of the loss Hessian is a good measure of sharpness for whether the training is linearly stable but not a good measure for generalization; ii) The common low-dimensional flatness-generalization perspective faces challenges in comprehending the high-dimensional loss landscape of neural networks. Generalization performance is a combined effect of most eigendirections, including those associated with small eigenvalues.

\section{Loss spike, $\lambda_{\text{max}}$ and condensation}

\subsection{Loss spike experimentally facilitates condensation}

In this section, we study the effect of loss spikes on condensation, which may improve the model’s generalization in some situations \citep{he2019control,jastrzkebski2017three}. A condensed network, which refers to a network with neurons condensing in several discrete directions, is equivalent to another smaller network \citep{zhou2021towards,luo2021phase}. It has a lower effective complexity than it appears. The embedding principle  \citep{zhang2021embedding,zhang2022embedding, fukumizu2019semi,simsek2021geometry} shows that a condensed network, although equivalent to a smaller one in approximation, has more degeneracy and descent directions that may accelerate the training process. The low effective complexity and simple training process may be underlying reasons for good generalization. We show that the loss spike can facilitate the condensation phenomenon for the noise-free full-batch gradient descent algorithm.

\begin{figure}[h]
	\centering
        \subfigure[ReLU]{\includegraphics[width=0.95\textwidth]{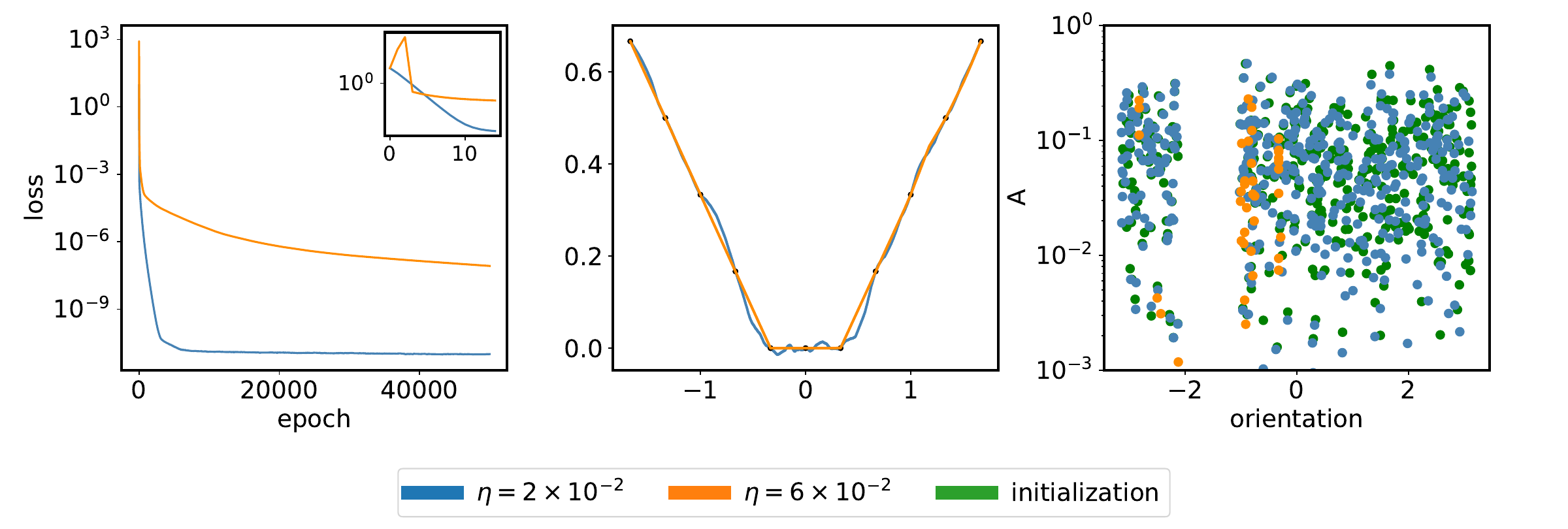}}
        \subfigure[Tanh]{\includegraphics[width=0.95\textwidth]{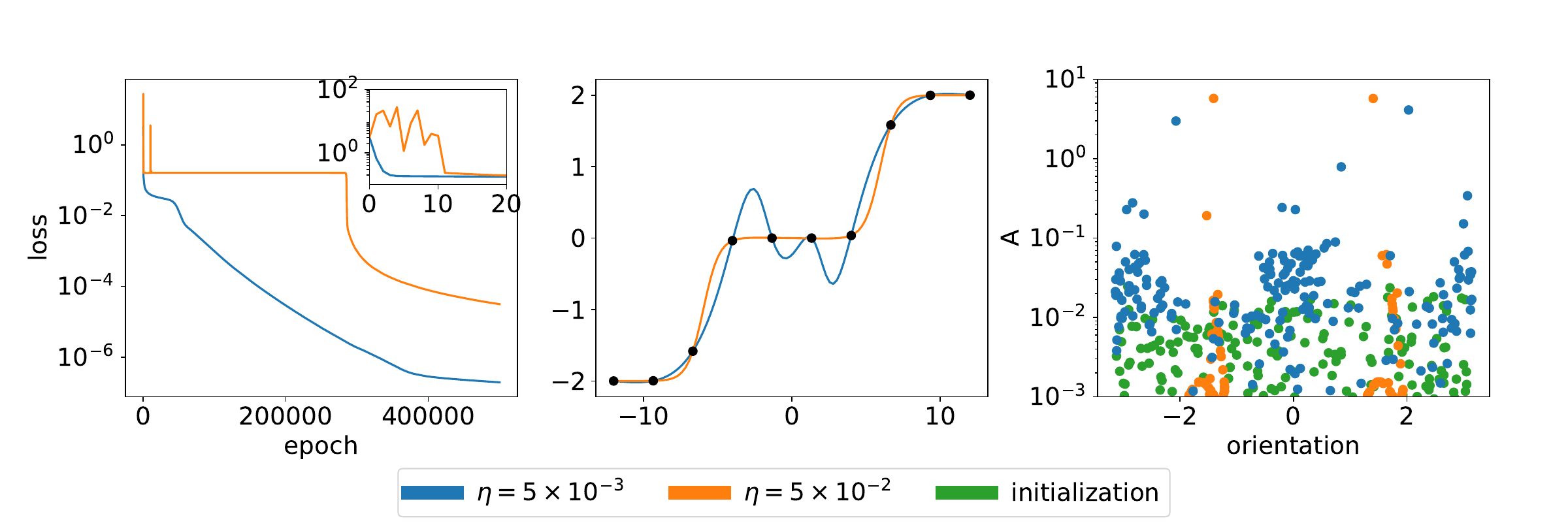}}

  \caption{Comparison of two-layer (a)ReLU (b)Tanh NNs with identical initialization but different learning rates $\eta$. The loss spike occurs at a large learning rate (orange), while not at a small learning rate (blue). Left: loss vs. epoch. The small picture in the upper right corner shows the occurrence of the loss spike in more detail. Middle: output. Right: The weight feature distribution of the trained models and the initial one.}
  \label{pic:condense}
\end{figure}

As shown in Fig. \ref{pic:condense}, we train a ReLU with 500 hidden neurons and a Tanh NN with 100 hidden neurons for the one-dimensional fitting problem to fit the data using MSE as the loss function. To clearly study the effect of loss spike on condensation, we take the parameter initialization distribution in the linear regime \citep{luo2021phase} that does not induce condensation without additional constraints. 
For NNs with identical initialization, we train the network separately with a small learning rate (blue) and a large learning rate (orange). For the left subfigure in Fig. \ref{pic:condense}, the loss value has a significant spike for the large learning rate, but not for the small one. At the same time, the middle subfigure reveals that the model output without a loss spike (blue) during the training process has more oscillation than the model output with a loss spike (orange). We study the features of parameters to understand the underlying effect of loss spike better.

To study the parameter features, we measure each parameter pair $(a_j,\vw_j)$ by the feature direction $\hat{\vw}_j=\vw_j/\norm{\vw_j}_{2}$ and amplitude \footnote{The amplitude accurately describes the contribution of ReLU neurons due to the homogeneity. For tanh neurons, there is a positive correlation between their amplitude and contribution. Appendix \ref{app:exp} provides a more refined characterization of tanh network features.} $A_j=|a_j|\norm{\vw_j}_{2}$. For a NN with one-dimensional input, after incorporating the bias term, $\vw_j$ is two-dimensional, and we use the angle between $\vw_j$ and the unit vector $(1,0)$ to indicate the orientation of each neuron. The scatter plots of $\{(\hat{\vw}_j,|a_j|)\}_{j=1}^{m}$ and $\{(\hat{\vw}_j,\norm{\vw_j}_2)\}_{j=1}^{m}$ of tanh activation are presented in Appendix \ref{app:exp} to eliminate the impact of the non-homogeneity of tanh activation.

The scatter plots of $\{(\hat{\vw}_j,A_j)\}_{j=1}^{m}$ of the NN is shown in the right subfigure of Fig. \ref{pic:condense}. Parameters without loss spikes (blue) are closer to the initial values (green) than those with loss spikes (orange). For the case with loss spikes, non-zero parameters tend to condense in several discrete orientations, showing a tendency to condensation.

Fig. \ref{pic:condense} shows that as the learning rate $\eta$ increases, the occurrence of loss spikes becomes more pronounced. At the same time, the constraints on $\lambda_{\text{max}}$ are also increasing, resulting in a decrease in the maximum eigenvalue $\lambda_{\text{max}}$ of the network. It is worth noting that this decrease is accompanied by an increase in the condensation level of network parameters, indicating a clear correlation between the two phenomena.

The variation in learning rates leads to the occurrence or absence of spikes, and we observe that this is accompanied by the model exhibiting either a condensed or non-condensed state. Therefore, it is crucial to investigate the relationship between the two under different learning rates to understand how these dynamics influence the model's training stability and parameter distribution.

\subsection{the correlation between $\lambda_{\text{max}}$ and the condensation}

The previous experiments shows that $\lambda_\mathrm{max}$ reaches a lower value during the spike process which might be the potential mechanism by which spikes promote condensation. We further wonder whether there is a correlation between $\lambda_{\text{max}}$ and the condensation. To quantitatively study this correlation, we define a measure to quantify the degree of condensation between neurons in a neural network:

\begin{definition}

Assuming we have a set containing $n$ two-dimensional data points, where each data point is represented as $(\theta_i, r_i)$ for $i = 1, 2, ..., n$. Each data point has two attributes: direction (represented by the angle of the vector) and magnitude (the length of the vector). Each point in our case can be obtained from a neuron parameter ($w_i, b_i$) in a specific hidden layer of the neural network:
$$\theta_i=arctan\frac{b_i}{w_i},\quad\quad r_i = \sqrt{w_i^2+b_i^2}.$$

The weighted average direction $\bar{\theta}$, where the weights are the magnitudes $r_i$ of each data point can be defined as:

\[
\bar{\theta} = \text{arctan}\left(\frac{\sum_{i=1}^{n} r_i \sin(\theta_i)}{\sum_{i=1}^{n} r_i},\frac{\sum_{i=1}^{n} r_i \cos(\theta_i)}{\sum_{i=1}^{n} r_i}\right)
,\]
where $\text{arctan}(y, x)$ is the arctangent function that returns the angle whose tangent is the quotient of its arguments $y$ and $x$.
\end{definition}

Next, using this weighted average direction $\bar{\theta}$, we calculate the variance of the weighted magnitudes.
\begin{definition}
We define the condensed variance is:
\[
\text{$\text{V}_{\text{cond}}$} = \frac{\sum_{i=1}^{n} r_i (\theta_i - \bar{\theta})^2}{\sum_{i=1}^{n} r_i}
.\]
\end{definition}

This definition is based on the condensation phenomenon we observed in experiments: the closer the neuron directions are, i.e., the smaller the variance, the higher the condensation level. At the same time, neurons with larger magnitudes should have greater weights. Based on this, we designed a weighted variance-based condensed variance. This provides the mathematical definitions for the degree of condensation between neurons in a neural network. Based on this, we conducted a series of experiments to investigate the correlation between $\lambda_\mathrm{max}$ and condensed variance.

\begin{figure}[h]
	\centering
	\includegraphics[width=0.4\textwidth]{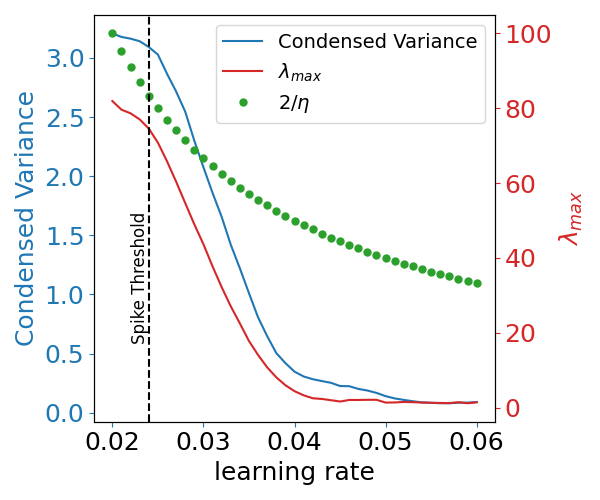}

  \caption{Condensed variance and $\lambda_{max}$ with varying learning rate. The condensed variance (blue) decreases as the learning rate increases. The $\lambda_{max}$ (red) value observed at the end of the training epoch exhibits a decreasing trend with respect to increasing learning rates, mirroring the behavior of the condensed variance. 2/$\eta$ (green) always remains higher than $\lambda_{max}$ (red). Learning rates greater than the threshold indicated by the vertical dashed line (black) will result in the observation of loss spikes. The training is conducted using ReLU NNs with the same settings as in Fig. \ref{pic:condense}}.
  \label{pic:condense_variance_lr}
\end{figure}

We make some observations using ReLU neural networks with identical settings to those employed in Fig. \ref{pic:condense}. 
We also conducted experiments (see Appendix \ref{app:act}) using other activation functions, such as Sigmoid and LeakyReLU, under identical settings.

As shown in Fig. \ref{pic:condense_variance_lr}. For varying learning rate, we observe condensed variance and $\lambda_{max}$ at the end of training in ReLU NNs. To more intuitively observe the threshold for the occurrence of spikes and condensation, we define the spike threshold. The spike threshold, determined through experimental observation, is marked in the figure by the vertical dashed line (black), indicates that when the learning rate exceeds this threshold, a clear observation of loss spikes \footnote{When the current loss is significantly greater than the loss from the previous epoch, it is considered a loss spike.} can be made. Also, the value of $\lambda_{max}$ (red) is always lower than 2/$\eta$ (green). We have observed similar trends in the changes of condensed variance and $\lambda_{max}$, which inspires us that the phenomenon of ‘‘loss spikes promoting condensation’’ may be related to $\lambda_{max}$. To further investigate the relationship between condensed variance and $\lambda_{max}$, we conducted experiments under different settings.

\begin{figure}[h!]
	\centering
	\includegraphics[width=0.8\textwidth]{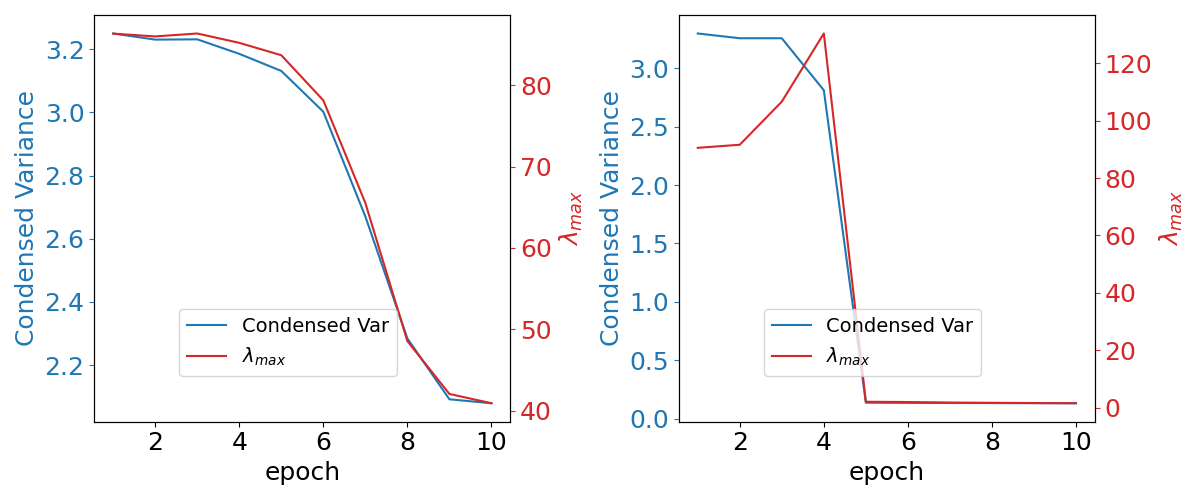}
  \caption{Condensed variance and $\lambda_{max}$ at the initial stages of training with learning rate = 0.03 (left) and 0.05 (right). The loss spike occurs at a large learning rate (right) but not a small learning rate (left). Similarly, the spikes in $\lambda_{max}$ (red) also follow this pattern, occurring at higher learning rates (right) but not at lower learning rates (left). Throughout the training process, $\lambda_{max}$ (red) consistently displays the same trend as the condensed variance (blue), irrespective of the occurrence of spikes. The training is conducted using ReLU NNs with the same settings as in Fig. \ref{pic:condense}}
  \label{pic:condense_variance_epoch}
\end{figure}

In our experimental setup, the changes in $\lambda_{max}$ and condensed variance occur primarily at the beginning of training and stabilize towards the end. Therefore, we study the trend of their values in the early stages of the training process. We investigate the behavior of condensed variance and $\lambda_{max}$ during the initial stages of training using learning rates of 0.03 and 0.05 under the neural network initialization settings of the linear regime \citep{luo2021phase} (Fig. \ref{pic:condense_variance_epoch}, left and right, respectively). Notably, a loss spike was observed when training with the high learning rate of 0.05 (right), but not with the low learning rate of 0.03 (left). Throughout the training process, whether or not loss spikes occurred, $\lambda_{max}$ (red) consistently exhibited the same trend as the condensed variance (blue), they both decrease during the same epoch (ignoring the temporary increase in $\lambda_{max}$.

\begin{remark}
We conduct the same experiments using the initialization settings of the condensed regime\citep{luo2021phase} (see Appendix \ref{app:condensed regime}).
\end{remark}

\begin{figure}[h!]
	\centering
	\includegraphics[width=0.4\textwidth]{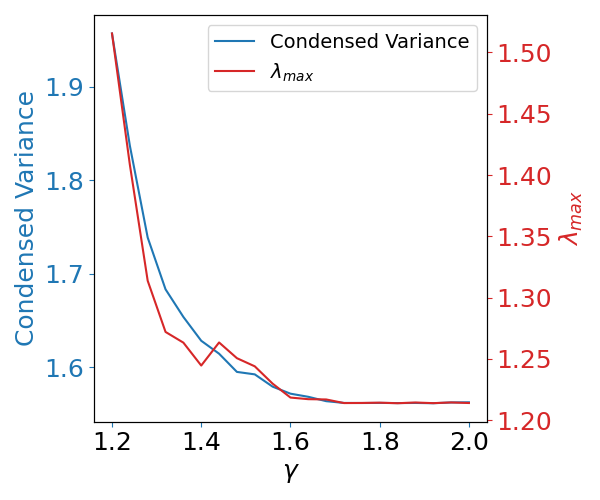}
  \caption{Condensed variance and $\lambda_{max}$ with varying $\gamma$ in condensed regime. The condensed variance (blue) decreases as $\gamma$ increases. The $\lambda_{max}$ (red) value observed at the end of the training epoch exhibits a decreasing trend with respect to increasing $\gamma$, mirroring the behavior of the condensed variance. The training is conducted using ReLU NNs with the same settings as in Fig. \ref{pic:condense}}
  \label{pic:condense_variance_gamma}
\end{figure} 

At the same time, we also investigated the trend changes of condensed variance and $\lambda_{max}$ under the neural network initialization settings of the condensed regime \citep{luo2021phase}. We found that, for different initialization settings, both still exhibit similar trends. By fixing the learning rate at 0.01 and taking 10 random seeds, we obtained the average values as shown in Fig. \ref{pic:condense_variance_gamma}, where $\gamma$ is defined in \citep{luo2021phase} (see Appendix \ref{app:setup}).

In our case, $\gamma > 1$ is the condensed regime. As the initialization variance decreases, both the condensed variance and $\lambda_{max}$ exhibit a similar decaying trend. In other words, at the end of the training phase, the network becomes more condensed and converges to a flatter loss landscape.

These experiments demonstrate a consistent correlation between condensed variance and $\lambda_{max}$ across various settings, strengthening our hypothesis that the loss spike phenomenon, which favors flatter minima (smaller $\lambda_{max}$), may play an important role in promoting weight condensation. The observed relationship between condensed variance and $\lambda_{max}$ provides valuable insights for understanding the underlying mechanisms of loss spikes and their impact on the learning dynamics of neural networks. Further theoretical and empirical investigations are needed to unravel the precise nature of this relationship and its implications for generalization performance.

\section{Conclusion and discussion}

In this work, we focus on loss spikes observed during neural network training and revisiting the relationship between flatness and generalization. We explain the ascent stage based on the landscape structure, specifically the LLAS structure. For the descent stage, we offer an explanation from the perspective of frequency analysis. We revisit the common understanding of the relationship between flatness and generalization through this frequency analysis. Additionally, we observe in our experiments that noise-free gradient descent with loss spikes can facilitate feature condensation, accompanied by flatter solutions, which may explain good generalization performance in some situations. We observe that there is a certain correlation between $\lambda_{max}$ and condensation, which might be the potential mechanism by which spikes promote condensation.

Clearly, many questions remain open. For example, why is the eigendirection corresponding to a large eigenvalue dominated by low-frequency components? Why can loss spikes facilitate feature condensation? We leave the discussion of these important questions for future work.

\section*{Acknowledgments}
This work is sponsored by the National Key R\&D Program of China Grant No. 2022YFA1008200, the National Natural Science Foundation of China Grant No. 92270001(Z. X.), 12371511 (Z. X.), 12422119 (ZX), Shanghai Municipal of Science and Technology Major Project No. 2021SHZDZX0102, and the HPC of School of Mathematical Sciences and the Student Innovation Center, and the Siyuan-1 cluster supported by the Center for High Performance Computing at Shanghai Jiao Tong University, Key Laboratory of Marine Intelligent Equipment and System, Ministry of Education, P.R. China. This work was partially supported by SJTU Kunpeng\&Ascend Center of Excellence.


\bibliographystyle{plainnat}  
\bibliography{bibfile}

\newpage
\appendix
\section{Experimental setups}\label{app:setup}

For Fig. \ref{pic:spike_real} (a-c), Fig. \ref{fig: LLAS}, Fig. \ref{pic:errorbar}, we use the two-layer tanh FNN with a width of 20 to fit the target function using full-batch gradient descent as follows, 

\begin{equation*}
    f(x)=\mathrm{sin}(x)+\mathrm{sin}(4x). 
\end{equation*}
The initialization of the parameters $\vtheta \sim N(0, m^{-1})$, where $m$ is the width of the NN, and the learning rate $\eta=0.05$. For Fig. \ref{pic:spike_real} (a), the $\lambda_{\rm max}$ is calculated every 100 epochs. Fig. \ref{pic:spike_real} (c) and Fig. \ref{fig: LLAS} show the parameter trajectories of different epoch intervals, which are indicated on the label of the color bar. For  Fig. \ref{pic:errorbar}, the $\vtheta_{\rm max}$ is selected at epoch 114320, and the $\vtheta_{\rm end}$ is selected at epoch 114400. 

For Fig. \ref{pic:spike_real} (d-f), we use the two-layer ReLU CNN with a Max Pooling layer behind the activation function for the CIFAR10-1k classification problem, i.e., using the first 1000 training data of the CIFAR10 as the training data. The number of the convolution kernels is 16 and the size is $3\times 3$. We use the MSE as the loss function with learning rate $\eta=0.1$. 

For Fig. \ref{pic:spike_real} (g-i), we use the VGG-11 \citep{simonyan2014very} model with batch normalization layer for the CIFAR10 classification problem. We conducted experiments with two different settings: a learning rate of 0.2 and another with a learning rate of 0.3. Furthermore, we performed PCA analysis on the spikes from the model trained with a learning rate of 0.3 to gain deeper insights into the spike dynamics.

For Fig.  \ref{pic:spike_toy}, we use the following quadratic model as the toy model to illustrate the LLAS structure, 
\begin{equation*}
    f(x,y)=(50 x+200)y^2-x+5,
\end{equation*}
where $(x,y) \in (-4, + \infty) \times \mathbb{R}$. The training uses the gradient descent algorithm with learning rate $\eta=5\times 10^{-3}$ and the initial value $(x, y)=(0.5, 0.00001)$. 

For Fig. \ref{pic:frequency} (a, b) and Fig. \ref{pic:projection} (a), we use the two-layer tanh FNN with a width of 500 to fit the target function using full-batch gradient descent as follows, 

\begin{equation*}
    f(x)=\mathrm{tanh}(x-6)+\mathrm{tanh}(x+6). 
\end{equation*}
The initialization of the parameters $\vtheta \sim N(0, m^{-0.4})$, where $m$ is the width of the NN, and the learning rate $\eta=0.001$. The training dataset is obtained by sampling 15 points equidistantly in the $[-12, 12]$ interval, and the test dataset is obtained by sampling 14 points equidistantly in the $[-11.14, 11.14]$ interval, which is approximately the midpoint of the pairwise data of the training set.

For Fig. \ref{pic:projection} (b-c), we use the CNNs for the CIFAR10-1k classification problem with structures shown in Table \ref{tab:picb}-\ref{tab:picc}, respectively. We use ReLU as the activation function, added behind each convolutional layer. We use the Xavier initialization and the MSE loss function. The learning rate is 0.005. For Fig. \ref{pic:projection} (d), we use the CNNs for the CIFAR10-2k classification problem with structures shown in Table \ref{tab:picd}. We use ReLU as the activation function, added behind each convolutional layer. We use the Xavier initialization and the cross-entropy loss function. The learning rate is 0.01. The large batch size we used is 1000, while the small one is 32. 

\begin{table}[h]
    \centering
    \caption{The architecture of the three-layer CNN used in Fig. \ref{pic:projection} (b).}
    \begin{tabular}{c|c}
    \hline \hline Layer & Output size \\
    \hline input & $32 \times 32 \times 3$ \\
    $3 \times 3 \times 16$, conv & $32 \times 32 \times 16$ \\
    $2 \times 2$, maxpool & $16 \times 16 \times 16$ \\
    \hline flatten & $4096$ \\
    $4096 \rightarrow 10$, linear & 10 \\
    \hline \hline
    \end{tabular}
    \label{tab:picb}
\end{table}

\begin{table}[h]
    \centering
    \caption{The architecture of the three-layer CNN used in Fig. \ref{pic:projection} (c).}
    \begin{tabular}{c|c}
    \hline \hline Layer & Output size \\
    \hline input & $32 \times 32 \times 3$ \\
    $3 \times 3 \times 16$, conv & $32 \times 32 \times 16$ \\
    $2 \times 2$, maxpool & $16 \times 16 \times 16$ \\
    $3 \times 3 \times 32$, conv & $16 \times 16 \times 32$ \\
    $2 \times 2$, maxpool & $8 \times 8 \times 32$ \\
    \hline flatten & $2048$ \\
    $2048 \rightarrow 10$, linear & 10 \\
    \hline \hline
    \end{tabular}
    \label{tab:picc}
\end{table}

\begin{table}[h!]
    \centering
    \caption{The architecture of the five-layer CNN used in Fig. \ref{pic:projection} (d).}
    \begin{tabular}{c|c}
    \hline \hline Layer & Output size \\
    \hline input & $32 \times 32 \times 3$ \\
    $3 \times 3 \times 16$, conv & $32 \times 32 \times 16$ \\
    $2 \times 2$, maxpool & $16 \times 16 \times 16$ \\
    $3 \times 3 \times 32$, conv & $16 \times 16 \times 32$ \\
    $2 \times 2$, maxpool & $8 \times 8 \times 32$ \\
    $3 \times 3 \times 64$, conv & $8 \times 8 \times 64$ \\
    $2 \times 2$, maxpool & $4 \times 4 \times 64$ \\
    \hline flatten & $1024$ \\
    $2048 \rightarrow 500$, linear & 500 \\
     $500 \rightarrow 10$, linear & 10 \\
    \hline \hline
    \end{tabular}
    \label{tab:picd}
\end{table}

For Fig. \ref{pic:condense} (b), Fig. \ref{fig:condense_tanh_further}, we use the two-layer tanh FNN with a width of 200 to fit the target function using full-batch gradient descent as follows, 
\begin{equation*}
    f(x)=\mathrm{tanh}(x-6)+\mathrm{tanh}(x+6). 
\end{equation*}
The initialization of the parameters $\vtheta \sim N(0, m^{-1})$, where $m$ is the width of the NN. We train the NN with loss spikes using the learning rate $\eta=0.05$ while using $\eta=0.005$ for the training without loss spikes. The training dataset is obtained by sampling 10 points equidistantly in the $[-12, 12]$ interval.

\begin{figure}[h!]
	\centering
	\includegraphics[width=0.8\textwidth]{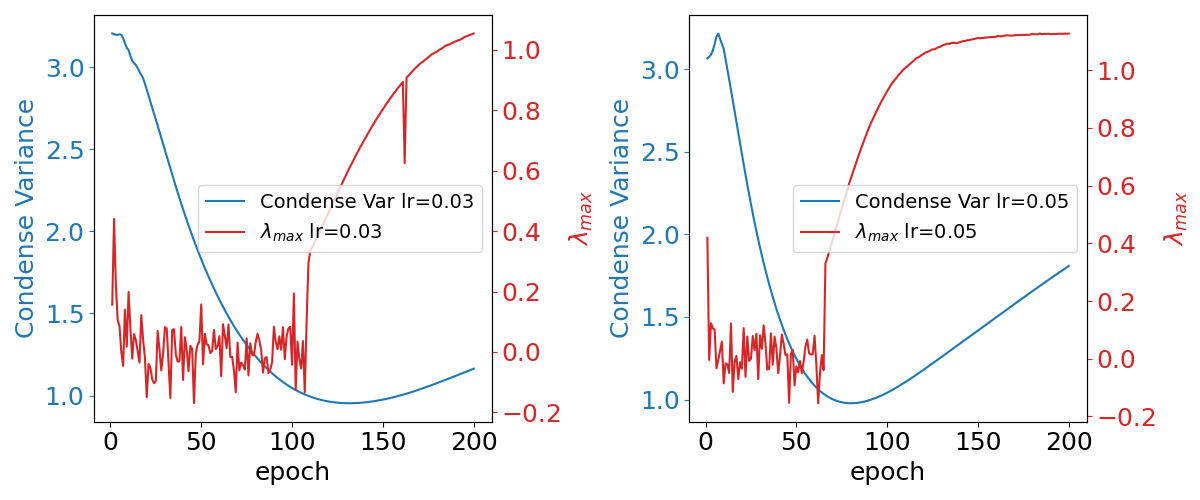}
  \caption{condensed variance and $\lambda_{max}$ at the initial stages of training with learning rate = 0.03 (left) and 0.05 (right).}
  \label{pic:condense_variance_epoch_condensedregime}
\end{figure}

For Fig. \ref{pic:condense_variance_epoch_condensedregime}, we use the two-layer ReLU FNN with a width of 500 to fit the target function using full-batch gradient descent as follows, 
\begin{equation*}
    f(x)=\frac{1}{2}\mathrm{ReLU}(-x-\frac{1}{3})+\frac{1}{2}\mathrm{ReLU}(x-\frac{1}{3}). 
\end{equation*}
The initialization of the parameters $\vtheta \sim N(0, m^{-0.8})$, where $m$ is the width of the NN. We train the NN with loss spikes using the learning rate $\eta=0.03$ and $\eta=0.05$  The training dataset is obtained by sampling 11 points equidistantly in the $[-5/3, 5/3]$ interval.

For Fig. \ref{pic:condense_variance_gamma}, we refer to a special version (with the same initialization for network parameters) of the definition of gamma in \citep{luo2021phase}:

We consider a two-layer NN with m hidden neurons:
$$f_{\boldsymbol{\theta}}(\boldsymbol{x})=\sum_{k=1}^m a_k \sigma\left(\boldsymbol{w}_k^{\top} \boldsymbol{x}\right),$$
where $\boldsymbol{x} \in \mathbb{R}^{\boldsymbol{d}}$, $\boldsymbol{\theta}=\left(\boldsymbol{\theta}_a, \boldsymbol{\theta}_{\boldsymbol{w}}\right)$ with $\boldsymbol{\theta}_a=\left(\left\{a_k\right\}_{k=1}^m\right)$, $\boldsymbol{\theta}_{\boldsymbol{w}}= \left(\left\{\boldsymbol{w}_k\right\}_{k=1}^m\right)$ is the set of parameters initialized by $a_k^0 \sim N\left(0, \beta_1^2\right), \boldsymbol{w}_k^0 \sim N\left(0, \beta_1^2 \boldsymbol{I}_d\right)$.

For the vanilla gradient flow training dynamics of NN, the hyperparameter $\beta_1$ is a function of $m$. To follow the analysis of initialization \citep{luo2021phase}, our initialization parameter is:
$$
\gamma=\lim _{m \rightarrow \infty}-\frac{\log \beta_1^{2}}{\log m},
$$

\newpage

\subsection{Detailed Features of LLAS structure} \label{app:LLAS_further}

We calculate the Hessian matrix and its eigenvalues as follows:

\begin{equation*}
    \frac{\partial^{2} f(x,y)}{\partial x^{2}}= 0. 
\end{equation*}
\begin{equation*}
    \frac{\partial^{2} f(x,y)}{\partial y^{2}}=100x+400 > 0. 
\end{equation*}
\begin{equation*}
    \frac{\partial^{2} f(x,y)}{\partial xy}=100y. 
\end{equation*}
\begin{equation*}
    \frac{\partial^{2} f(x,y)}{\partial yx}=100y. 
\end{equation*}

\[
\text{Hessian} = 
\begin{bmatrix}
\frac{\partial^{2} f(x,y)}{\partial x^{2}} & \frac{\partial^{2} f(x,y)}{\partial xy} \\
\frac{\partial^{2} f(x,y)}{\partial yx} & \frac{\partial^{2} f(x,y)}{\partial y^{2}}
\end{bmatrix} 
=
\begin{bmatrix}
0 & 100y \\
100y & 100x + 400
\end{bmatrix}.
\]

The eigenvalues of the Hessian matrix when $y=0$ are $100x+400>0$ and $0$.

\section{Experimental results}\label{app:exp}

\subsection{condensed variance and $\lambda_{max}$ in condensed regime} \label{app:condensed regime}

We investigated the behavior of condensed variance and $\lambda_{max}$ during the initial stages of training using learning rates of 0.03 and 0.05 under the neural network initialization settings of the condensed regime \citep{luo2021phase}.

However, the results do not display similar trends, but it is reasonable. When neural network parameters are initialized with extremely small values in the condensed regime, $\lambda_{max}$ starts off very small and even close to 0. After a period of training, $\lambda_{max}$ increases to a non-zero constant. This increase is necessarily accompanied by a rise in $\lambda_{max}$. Meanwhile, the degree of condensation of the neural network increases i.e., the condensed variance decreases since it is under the condensed regime. There is bound to be an inconsistency in the trends here. On the other hand, small initialization implies that $\lambda_{max}$ is very small and will not be affected by $2/\eta$, so overall, it may not have a strong correlation with condensation at the beginning of training under the neural network initialization settings of the condensed regime.

\subsection{Detailed Features of Tanh NNs} \label{app:tanh_further}

\begin{figure}[h]
	\centering
	\subfigure[initialization]{\includegraphics[width=0.32\textwidth]{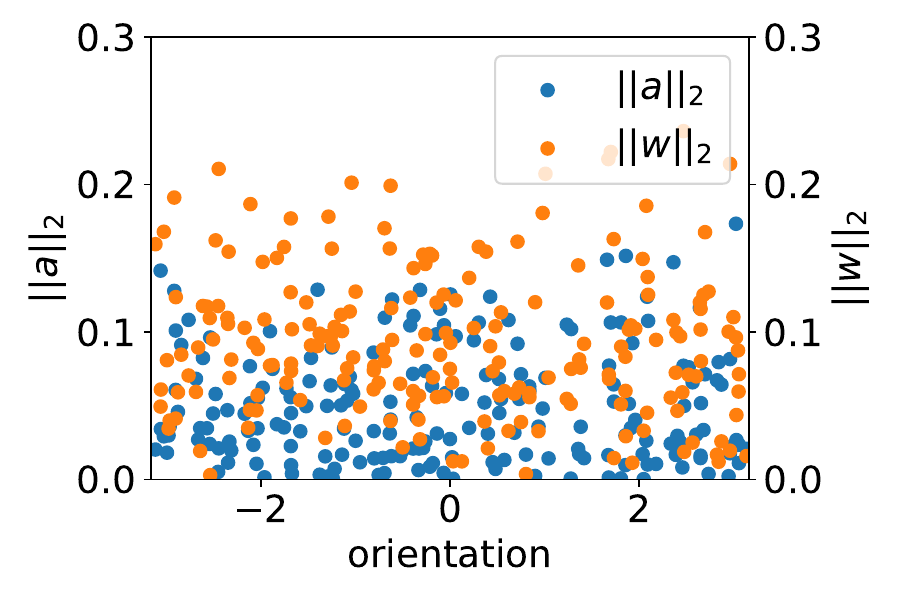}}
         \subfigure[$\eta=5\times10^{-3}$]{\includegraphics[width=0.32\textwidth]{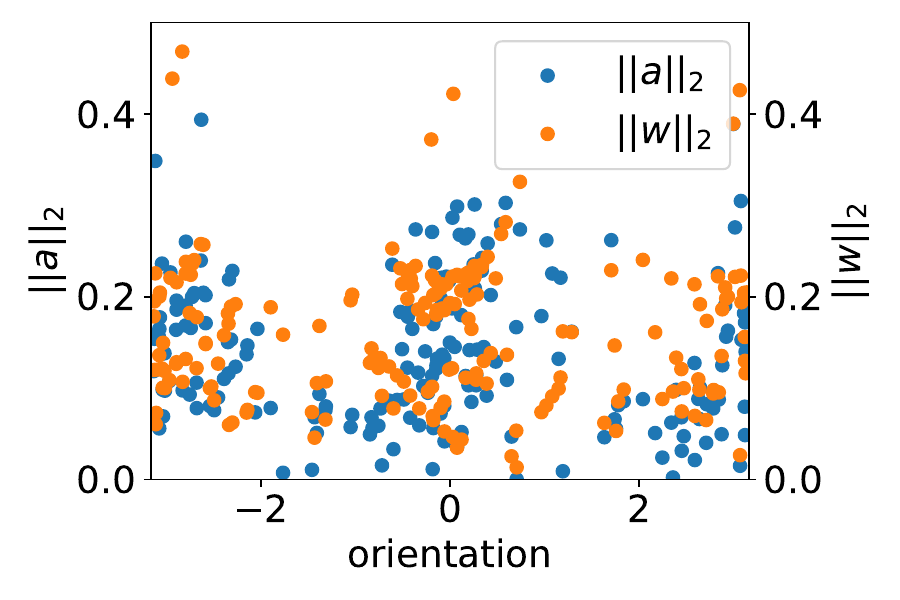}} 
         \subfigure[$\eta=5\times10^{-2}$]{\includegraphics[width=0.32\textwidth]{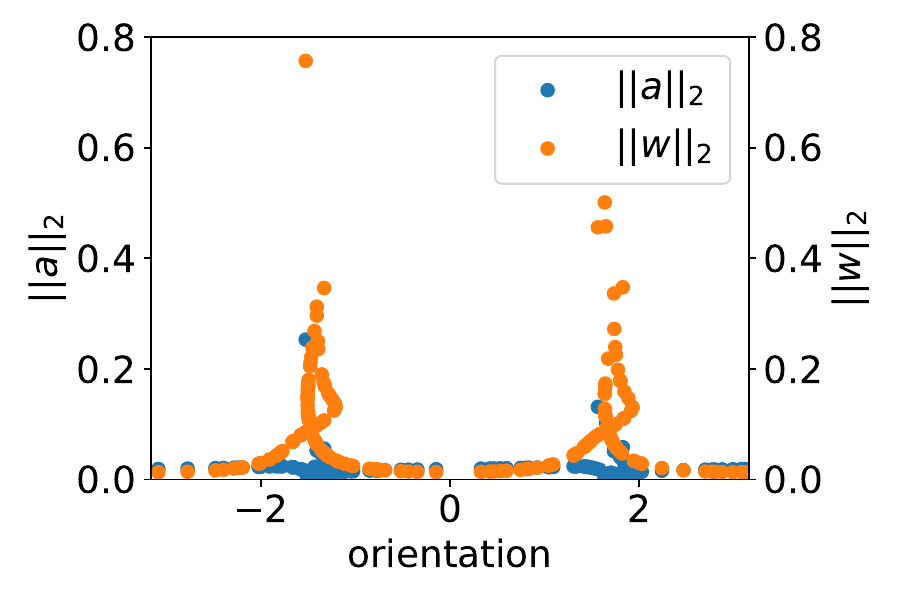}} 

  \caption{The normalized scatter diagrams between $\norm{a_j}$, $\norm{\vw_j}$ and the orientation of tanh NNs for the initialization parameters and the parameters trained with and without loss spikes. Blue dots and orange dots are the output weight distribution and the input weight distribution, respectively.} 
  \label{fig:condense_tanh_further}
\end{figure}

In order to eliminate the influence of the inhomogeneity of the tanh activation function on the parameter features of Fig. \ref{pic:condense} (b), we plot the normalized scatter figures between $\norm{a_j}$, $\norm{\vw_j}$ and the orientation, as shown in Fig. \ref{fig:condense_tanh_further}. Obviously, for the network with loss spikes, both the input weight and the output weight have weight condensation, while the network without loss spikes does not have weight condensation.

\subsection{Experimental results with different activation functions} \label{app:act}

\begin{figure}[h]
	\centering
	\includegraphics[width=0.4\textwidth]{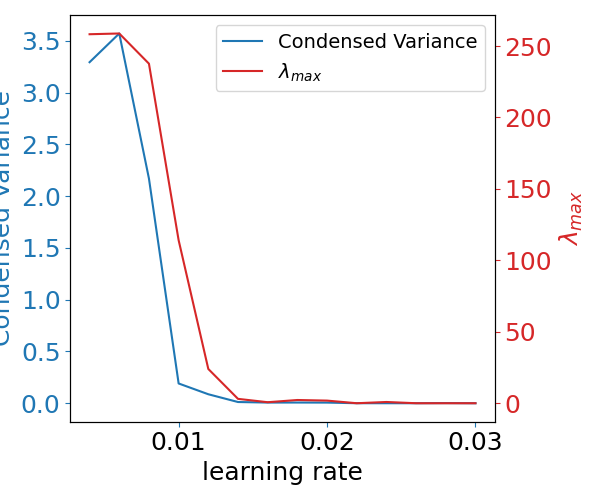}
  \caption{Condensed variance and $\lambda_{max}$ with varying learning rate. The condensed variance (blue) decreases as the learning rate increases. The $\lambda_{max}$ (red) value observed at the end of the training epoch exhibits a decreasing trend with respect to increasing learning rates, mirroring the behavior of the condensed variance. The training is conducted using Sigmoid NNs with the same settings as in Fig. \ref{pic:condense}}
  \label{fig:Sigmoid}
\end{figure}

\begin{figure}[h]
	\centering
	\includegraphics[width=0.8\textwidth]{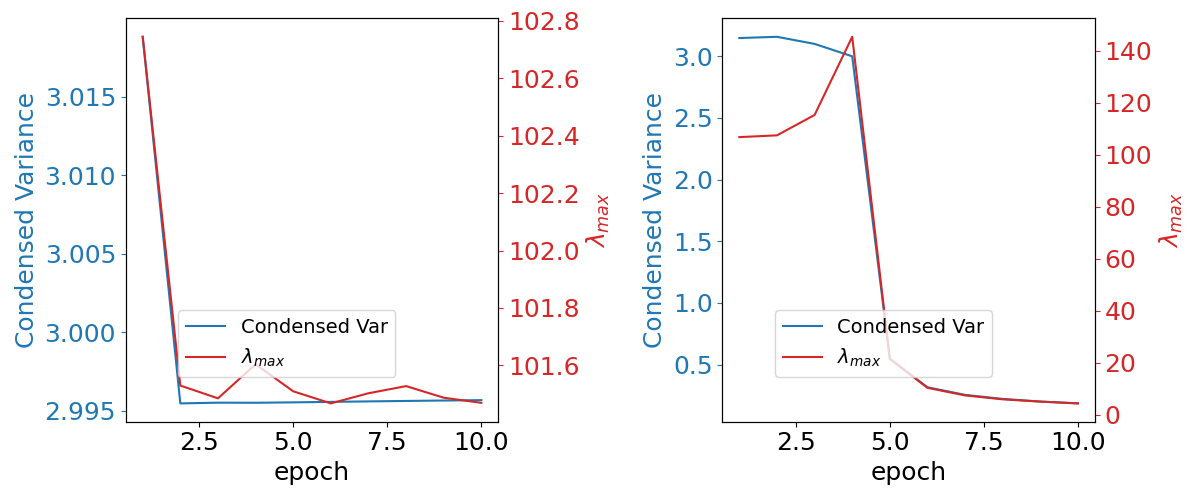}
  \caption{Condensed variance and $\lambda_{max}$ at the initial stages of training with learning rate = 0.01 (left) and 0.04 (right). Throughout the training process, $\lambda_{max}$ (red) consistently displays the same trend as the condensed variance (blue), irrespective of the occurrence of spikes. The training is conducted using LeakyReLU NNs with the same settings as in Fig. \ref{pic:condense}}
  \label{fig:leakyReLU}
\end{figure} 

We observed the phenomena shown in Fig. \ref{pic:condense_variance_lr} and Fig. \ref{pic:condense_variance_epoch} under the experimental setup, conducting experiments with both the Sigmoid activation function and leakyReLU. All other experimental settings were consistent with those in Fig. \ref{pic:condense_variance_lr} and Fig. \ref{pic:condense_variance_epoch}. The results obtained are as Fig. \ref{fig:Sigmoid} and Fig. \ref{fig:leakyReLU}.

\end{document}